%% file: arxiv.tex
\documentclass[10pt,twocolumn,letterpaper]{article}

\usepackage[pagenumbers]{iccv}   
\usepackage[accsupp]{axessibility}  
\usepackage{array}
\usepackage{multirow}
\usepackage{makecell}
\usepackage{pifont}
\usepackage{xcolor}
\usepackage{enumitem}
\usepackage{caption}
\usepackage{listings}
\usepackage{xcolor}
\usepackage{colortbl}
\usepackage{courier} %
\lstdefinestyle{singlecolumnstyle}{
  basicstyle=\ttfamily\scriptsize\fontfamily{pcr}\selectfont, 
  backgroundcolor=\color[HTML]{f8f8f8}, 
  frame=single,
  rulecolor=\color{gray},
  breaklines=true,
  breakindent=0pt
}
\lstdefinestyle{doublecolumnstyle}{
  basicstyle=\ttfamily\scriptsize\fontfamily{pcr}\selectfont, 
  backgroundcolor=\color[HTML]{f8f8f8}, %
  frame=single,
  rulecolor=\color{gray},
  breaklines=true,
  breakindent=0pt
}

\newcommand{\cmark}{\textcolor{black}{\ding{51}}}%
\newcommand{\xmark}{\textcolor{black}{\ding{55}}}%
\definecolor{iccvblue}{rgb}{0.21,0.49,0.74}
\usepackage[pagebackref,breaklinks,colorlinks,allcolors=iccvblue]{hyperref}

\def\paperID{7228} 
\def\confName{ICCV}
\def\confYear{2025}

\title{\textit{CATP-LLM}: Empowering Large Language Models for Cost-Aware Tool Planning}

\author{
Duo Wu\textsuperscript{1},
Jinghe Wang\textsuperscript{1}, 
Yuan Meng\textsuperscript{2}, 
Yanning Zhang\textsuperscript{1}, 
Le Sun\textsuperscript{1}, 
Zhi Wang\textsuperscript{1}
\\
\textsuperscript{1}Shenzhen International Graduate School, Tsinghua University\\
\textsuperscript{2}Department of Computer Science, Tsinghua University 
}

\newcommand{\ICCVRevision}[1]{ \textcolor{black}{#1}}

\begin{document}

\twocolumn[{%
\renewcommand\twocolumn[1][]{#1}%
\maketitle
\begin{center}
    \vskip -0.18in
    \includegraphics[width=0.93\textwidth]{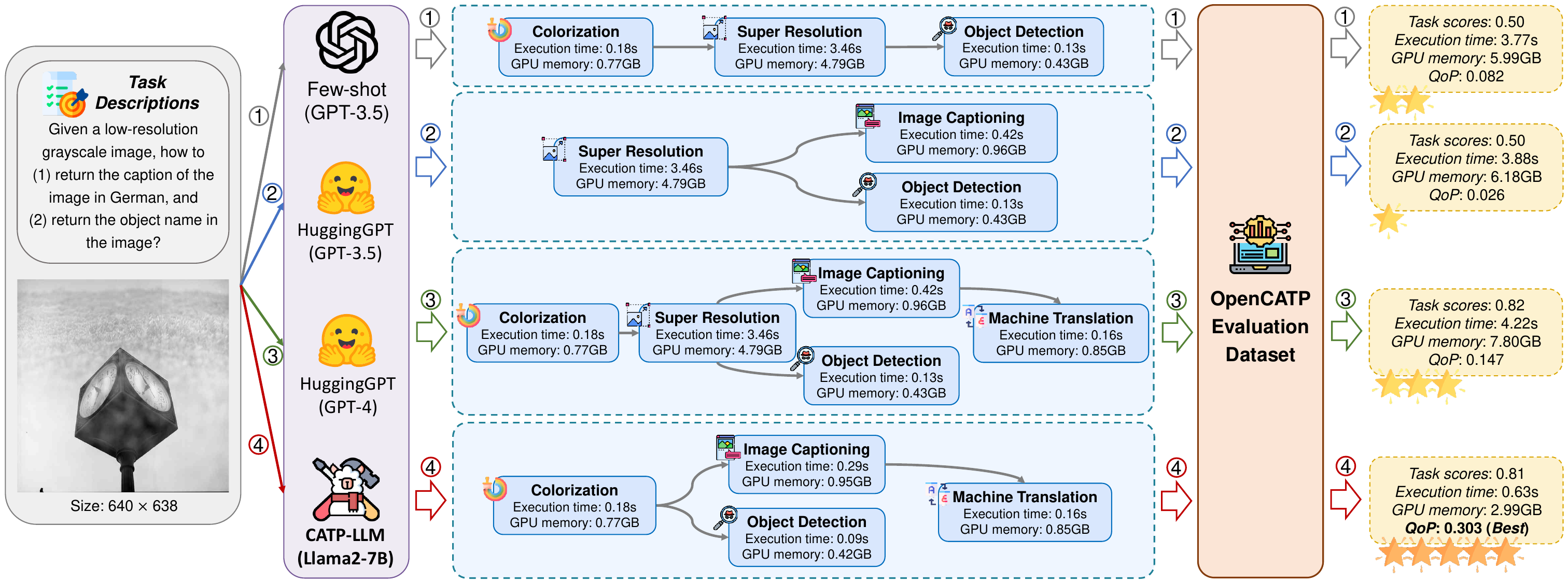}
    \vspace{-0.2cm}
    \captionof{figure}{We introduce \textit{CATP-LLM}, an efficient framework that enables LLMs to create logically complex tool plans for task solving while striking a good balance between plan performance and tool execution costs. We further present \textit{OpenCATP}, a cost-aware evaluation dataset with the Quality of Plan (QoP) metric to quantitatively evaluate a plan based on its task performance and execution costs (\eg, time and memory consumption). We demonstrate that CATP-LLM with the small Llama2-7B as backbone significantly outperforms GPT-powered methods in terms of higher QoP, with significantly lower execution costs and higher or comparable task performance. }
    \label{fig:example_figure}
 \end{center}%
}]

    

\begin{abstract}
\vskip -0.1in

Utilizing large language models (LLMs) for tool planning has emerged as a promising avenue for developing general AI systems, where LLMs automatically schedule external tools (e.g.,\;vision models) to tackle complex tasks based on task descriptions. To push this paradigm toward practical applications, it is crucial for LLMs to consider tool execution costs (e.g.,\;execution time) for tool planning. Unfortunately, prior studies overlook the tool execution costs, leading to the generation of expensive plans whose costs outweigh their benefits in terms of task performance. To fill this gap, we propose the \underline{C}ost-\underline{A}ware \underline{T}ool \underline{P}lanning with \underline{LLM}s (CATP-LLM) framework, which for the first time provides a coherent design to empower LLMs for cost-aware tool planning. 
Specifically, To facilitate efficient concurrent tool execution and cost reduction, we design a tool planning language to enhance the LLM for creating multi-branch non-sequential plans.
Moreover, we propose a cost-aware offline reinforcement learning algorithm to fine-tune the LLM to optimize the performance-cost trade-off in tool planning. In the lack of public cost-related datasets, we further present OpenCATP, the first dataset for cost-aware planning, which comprises 11,100 evaluation samples from diverse tasks. 
Extensive experiments show that CATP-LLM outperforms GPT-4 even when using Llama2-7B as its backbone, with the average improvement of 1.5\%-93.9\% in terms of plan quality.
Codes and dataset are available at: \url{https://github.com/duowuyms/OpenCATP-LLM}.
\end{abstract}
\vspace{-0.4cm}

\section{Introduction}
\label{sec:introduction}
Recent advances in large language models (LLMs)~\cite{achiam2023gpt,glm2024chatglm} have highlighted their extraordinary ability in tool planning~\cite{schick2024toolformer,yang2024gpt4tools,gao2024clova}, where LLMs can schedule and execute external tools (\eg, vision models~\cite{li2024formal,shen2024hugginggpt} and Python programs~\cite{gupta2023visual,hu2024visual}) to tackle complex tasks based on task descriptions. 
For instance, when confronted with the task ``\textit{How to generate the caption of a low-resolution image in German}'', the LLM may devise the following tool plan: (1) use a super resolution tool to enlarge the image; (2) employ the image captioning model to generate caption; (3) execute a translator to convert the caption into German. 
Due to the flexibility of tool usage and composition, this paradigm has shown exceptional performance, especially in solving complex tasks
including visual question answering~\cite{suris2023vipergpt,yang2022empirical}, image editing~\cite{yang2023mm,wu2023visual,liu2023interngpt} and mathematical reasoning~\cite{zhang2023evaluating,wu2024mathchat}, paving the way for the development of  more general AI systems~\cite{gupta2023visual,ge2024openagi}.
Furthermore, it offers significant practical prospects by effectively utilizing existing tools and models, thereby reducing the requirement for developing and training new models~\cite{shen2024hugginggpt}.

Despite its promising potential, using LLMs for tool planning still faces practical issues related to plan execution costs. 
Specifically, the execution of tools will inevitably cause computation costs such as execution time and memory consumption, as depicted in Figure~\ref{fig:example_figure}.
This makes it essential for LLMs to carefully take into account tool execution overhead  when devising tool plans for specific tasks. 
Otherwise, they may generate expensive tool plans whose execution costs outweigh task performance, ultimately hindering their practical applications (\eg, Figure~\ref{fig:example_figure}\ding{174}).

However, empowering LLMs for cost-aware tool planning poses the following challenges.
\begin{itemize}
    \item First, \textit{how to enable LLMs for non-sequential tool planning for potential cost reduction?} 
    Non-sequential planning refers to the capability to create logically nonlinear plans that comprise multiple branches for concurrent tool execution. 
    It not only facilitates more efficient parallel processing of tools to reduce execution costs (\eg, execution time), but also enhances the ability of LLMs to solve complicated tasks that require complex tool plans~\cite{ge2024openagi}, as exemplified in Figure~\ref{fig:example_figure}\ding{175}. 
    Nevertheless, implementing non-sequential planning remains a challenge as recent insights have revealed that it can be difficult for LLMs to understand and generate logically nonlinear content through natural language prompts~\cite{huang2024can,perozzi2024let,yao2024open,beurer2024guiding}. 
    \item Second,  \textit{how to enable LLMs to achieve the optimal performance-cost trade-off in tool planning?} The key obstacle lies in the conflicts between plan performance and execution costs.
    Generally, simple plans often incur lower costs but may compromise performance, while complex plans can enhance performance but at the expense of higher costs. In addition, the costs of executing tool plans can be affected by the input data associated with the task (\eg, larger images increase object detection overhead). 
    Consequently, striking a good balance between performance and costs in tool planning remains a significant challenge for LLMs, which requires a deep understanding of the performance-cost trade-off and the impacts of input data on execution costs.
    \item  Finally, \textit{in the lack of public datasets that consider non-sequential planning and  tool execution costs, how to evaluate the  efficacy of LLMs in cost-aware tool planning?}
\end{itemize}

Prior studies~\cite{li2024formal,lu2024chameleon,shen2024hugginggpt,ge2024openagi} fail to achieve cost-aware tool planning, as most of them only support sequential planning and overlook the tool execution costs. 
To fill this research gap, in this paper, we propose the \underline{C}ost-\underline{A}ware \underline{T}ool \underline{P}lanning with \underline{LLM}s (CATP-LLM) framework, which for the first time provides a coherent design to efficiently enable LLMs for cost-aware tool planning.
Specifically, we propose the following techniques to systematically overcome the aforementioned challenges:
\begin{itemize}
    \item First, we design a tool planning language (TPL) to enhance the non-sequential planning capability of LLMs. 
    TPL transforms tools and their dependencies into learnable tokens, which, like language tokens, can be trained to learn the optimal representations for LLMs to understand their semantic information.
    TPL then encodes any tool plan, including non-sequential ones, as a sequence of tool and dependency tokens, so that the LLMs can generate diverse tool plans simply by  predicting these tokens. Additionally, it also introduces some special tokens to describe the structural information of tool plans to facilitate LLMs to generate tool plans of complex  structures.
    \item Based on the TPL, we then propose an efficient cost-aware offline reinforcement learning (CAORL) algorithm to address the second challenge. Specifically, CAORL introduces a context augmentation scheme for LLMs to capture the impacts of input data on the tool execution costs for effective cost-aware planning. It then fine-tunes LLMs based on the efficient offline RL algorithm~\cite{agarwal2020optimistic,prudencio2023survey,chen2021decision} with a dedicated reward model to guide the LLM to minimize plan execution costs while maximizing plan performance. 
    Besides, a data generation method is also designed to generate the training data for the LLM fine-tuning.
    \item To comprehensively assess the efficacy of LLMs in cost-aware tool planning, we establish a new dataset OpenCATP. 
    The key features of OpenCATP lie in three aspects: (1) it allows the measurement of plan costs; (2) it incorporates a unified metric Quality of Plan (QoP) to quantitatively evaluate the plan quality based on its performance and execution costs; (3) it comprises 11,100 samples including 2,400 samples for the non-sequential planning evaluation (\eg, Figure~\ref{fig:example_figure}).
\end{itemize}

We compare CATP-LLM with existing methods on OpenCATP. 
Results demonstrate that CATP-LLM outperforms GPT-3.5~\cite{chatgpt} and GPT-4~\cite{achiam2023gpt} with Llama2-7B~\cite{touvron2023llama} as the backbone LLM. To be specific, it achieves 1.02x-1.41x and 1.92x-5.99x higher QoP in the case of sequential planning and non-sequential planning, respectively, with lower execution costs and higher or comparable plan performance.

In summary, our contributions are three-fold:
\begin{itemize}
    \item We present CATP-LLM, the first framework that efficiently enables LLMs for cost-aware tool planning.
    \item We design the TPL language to enhance the LLMs for non-sequential planning, and propose the CAORL algorithm to efficiently fine-tune LLMs to strike  a good balance between performance and costs in tool planning.
    \item We develop a new dataset OpenCATP for the research community to comprehensively assess the performance of LLMs in cost-aware tool planning.
\end{itemize}

\section{Related Work}
\subsection{Tool Planning with LLMs}

\begin{table}[t]
    \centering
    \scriptsize
    \begin{tabular}{@{\;}m{2cm}<{\centering}@{\;}|@{\;\;} m{1.1cm}<{\centering} @{} m{1.45cm}<{\centering} @{}m{1.6cm}<{\centering}@{} m{1.45cm}<{\centering}@{} }
         \toprule
         \textbf{Method} & \textbf{Paradigm} & \textbf{Massive Tools}  & \textbf{Non-sequential Planning} & \textbf{Cost Awareness} \\
         \midrule
         VISPROG~\cite{gupta2023visual} & Prompt & \cmark  & \xmark & \xmark \\
         ART~\cite{paranjape2023art} & Prompt & \cmark  & \xmark & \xmark \\
         Chameleon~\cite{lu2024chameleon} & Prompt & \cmark & \xmark & \xmark \\
         HYDRA~\cite{ke2024hydra} & Prompt & \cmark & Limited & \xmark \\
         Formal-LLM~\cite{li2024formal} & Prompt & \cmark&  Limited & \xmark \\
         ControlLLM~\cite{liu2023controlllm} & Prompt & \cmark & Limited & \xmark \\
         HuggingGPT~\cite{shen2024hugginggpt} & Prompt & \cmark& Limited & \xmark \\
         \midrule
         ToolFormer~\cite{schick2024toolformer} & Fine-tune & \xmark& \xmark & \xmark \\
         GPT4Tools~\cite{yang2024gpt4tools} & Fine-tune & \xmark & \xmark & \xmark \\
         ToolkenGPT~\cite{hao2024toolkengpt} & Fine-tune & \cmark & \xmark & \xmark \\
         TRICE~\cite{qiao2023making} & RL & \cmark & \xmark & \xmark \\
         RLTF~\cite{ge2024openagi} & RL & \cmark & \xmark & \xmark \\
         \midrule
         CATP-LLM (\textbf{Ours}) & RL & \cmark & \cmark & \cmark \\
         \bottomrule
    \end{tabular}
    \vspace{-0.3cm}
    \caption{Comparison between our work with representative LLM-based tool planning methods.}
    \label{table:related_work_cmp}
    \vspace{-0.4cm}
\end{table}

Large language models (LLMs)~\cite{glm2024chatglm,touvron2023llama} have shown exceptional performance in understanding task descriptions  and deriving tool execution plans based on the provided toolkit to address the target tasks~\cite{gao2023assistgpt,liu2023controlllm,suris2023vipergpt,liu2023llava}. 
Therefore, tool planning with LLMs has emerged as a promising avenue to develop general AI systems~\cite{gupta2023visual,ge2024openagi}. In pursuit of this vision, numerous works have been proposed to enhance the tool planning abilities of LLMs, which can be categorized into two paradigms. 
The first paradigm leverages prompt engineering~\cite{chen2023unleashing,sahoo2024systematic,liu2023pre} to guide LLMs for tool scheduling, which has recently been extended by adding in-context demonstrations~\cite{yang2023mm,paranjape2023art,suris2023vipergpt,gupta2023visual}, introducing more tools~\cite{lu2024chameleon,wu2023visual}, or designing more dedicated planning procedures~\cite{shen2024hugginggpt,liu2023controlllm,li2024formal,ke2024hydra}.
Another line of work involves fine-tuning LLMs for tool planning through instruction tuning~\cite{yang2024gpt4tools,schick2024toolformer,hao2024toolkengpt,patil2023gorilla}. Recent advancements have also introduced reinforcement learning techniques~\cite{ge2024openagi,qiao2023making} to further augment LLMs for tool planning.


In Table~\ref{table:related_work_cmp}, we compare CATP-LLM with other LLM-based tool planning methods. Existing methods exhibit limited support for non-sequential planning, restricting the full potential of LLMs in cost-aware tool planning.
For instance, prior work~\cite{li2024formal} combines formal language with natural language to improve the planning ability of the LLM, but it necessitates domain experts to handcraft specialized prompts, which can be labor-intensive.
Although methods proposed in~\cite{shen2024hugginggpt,liu2023controlllm} integrate non-sequential plans  in JSON-like format into prompts as demonstrations, recent insights~\cite{huang2024can,perozzi2024let,shorten2024structuredrag,yao2024open} suggest that LLMs may not effectively understand and generate logically nonlinear information through language instructions, thereby hindering these methods to function effectively. 
In contrast, CATP-LLM incorporates a tool planning language to empower LLMs for non-sequential planning to effectively generate tool plans of various structures. Moreover, while existing methods overlook tool  execution costs, CATP-LLM introduces an efficient  algorithm to fine-tune LLMs for cost optimization.

\subsection{Reinforcement Learning for Tool Planning}

Reinforcement learning (RL)~\cite{wang2022deep,qiao2023making} has been proposed to enhance LLMs for tool planning, which enables LLMs to understand their actions and adjust behavior accordingly by  replaying plan execution feedback to  LLMs. 
However, existing methods~\cite{ge2024openagi,qiao2023making} suffer from coarse-grained planning, as they replay feedback to LLMs only at the completion of plan execution without any intermediate guiding signals.
In contrast, CATP-LLM introduces an efficient RL algorithm tailored for cost-aware tool planning. It allows LLMs to access  the plan quality each time a decision is made (\eg, adding a new tool to the plan). 
Such fine-grained perception empowers LLMs for more effective  cost-aware planning (\eg, selecting a cheaper tool if the current plan costs are already excessive).

\subsection{Datasets for Tool Planning}
With the advancement of tool planning with LLMs, numerous datasets have been established to access LLMs' capability in scheduling tools for task solving~\cite{wang2024gta,shen2023taskbench,lu2024toolsandbox,zhuang2023toolqa,ge2024openagi}. 
\ICCVRevision{However, existing datasets overlook non-sequential planning and tool execution costs, the crucial aspects that significantly affect the efficacy of tool planning in real-world scenarios. 
To bridge this gap, this work develops a new dataset OpenCATP that for the first time introduces many challenging tasks for non-sequential planning evaluation and a comprehensive scheme to evaluate plan quality based on its task performance and execution costs.}

\begin{figure*}[t]
    \centering
    \includegraphics[width=0.89\textwidth]{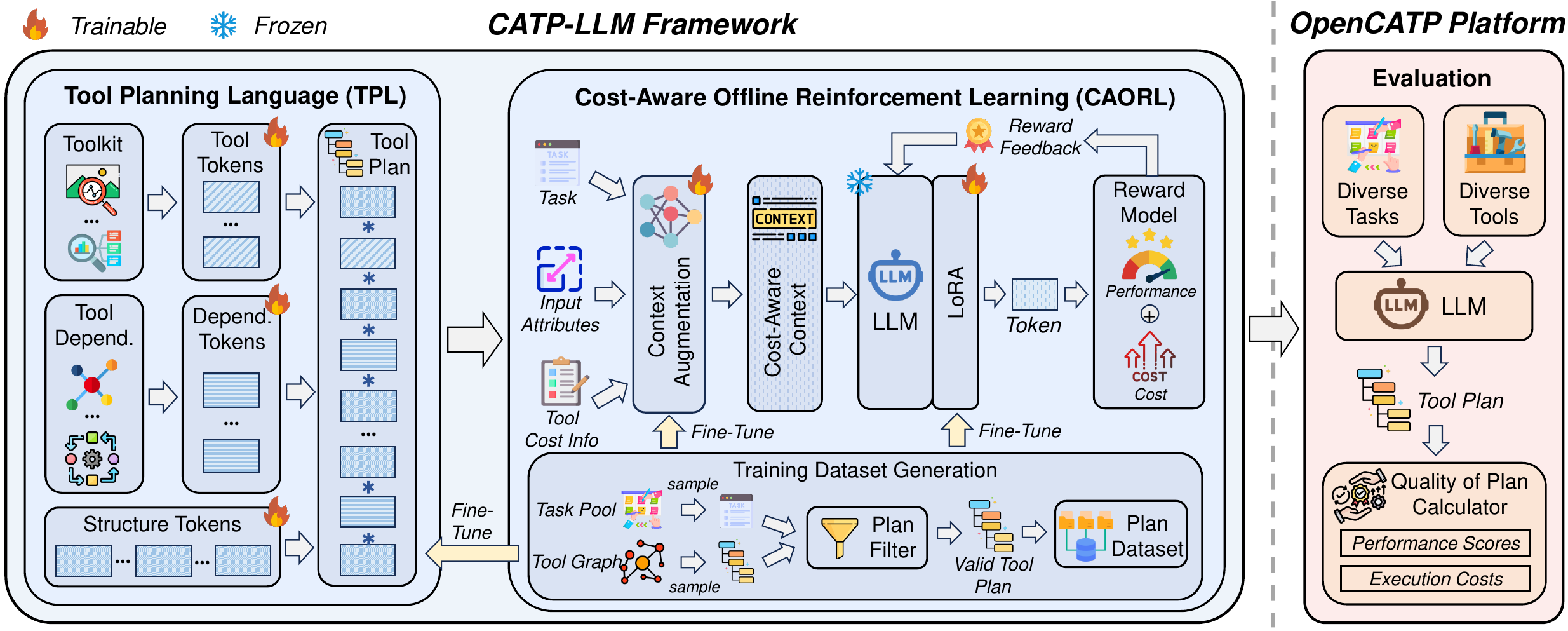}
    \vspace{-0.25cm}
    \caption{CATP-LLM consists of two core components: (1) a tool planning language to encode any tool plan as a sequence of trainable tokens for efficient tool planning, and (2) a cost-aware offline reinforcement learning algorithm to fine-tune the LLM for plan token generation to optimize the performance-cost trade-off. OpenCATP features diverse tools and tasks as well as a Quality of Plan (QoP) calculation mechanism to comprehensively evaluate the  quality of a plan based on its task performance and execution costs.}
    \vspace{-0.3cm}
    \label{fig:framework}
\end{figure*}

\section{CATP-LLM Design}
\label{sec:design}
We propose CATP-LLM, a general model-agnostic framework that enhances the capability of LLMs in cost-aware tool planning. As shown in Figure~\ref{fig:framework}, CATP-LLM consists of two core design building blocks:
\begin{itemize}
    \item \textbf{Tool planning language (TPL).} TPL encodes tools and their dependencies into trainable tokens, and represents any tool plan as a sequence of tokens augmented with special tokens to capture the structural information of the plan ($\S$\ref{subsec:tool_planning_language}).
    \item \textbf{Cost-aware offline reinforcement learning (CAORL).} Built on TPL, CAORL fine-tunes the LLM for plan token generation to optimize the trade-off between plan performance and costs ($\S$\ref{subsec:caorl}). 
\end{itemize}

\subsection{Tool Planning Language}
\label{subsec:tool_planning_language}
Given a set of tools $\mathcal{T}=\{t_1, t_2, \cdots\}$ and their dependencies $\mathcal{D}=\{d_1, d_2, \cdots\}$, a tool plan can be expressed as a directed acyclic graph (DAG) $p=(\{t_i\}_{i=1}^n, \{d_j\}_{j=1}^m)$, where node $t_i\in\mathcal{T}$ indicates a specific tool and edge $d_j\in\mathcal{D}$ denotes a specific dependency (\eg, a resource dependency where one tool accepts the outputs of another tool as inputs). To enable the LLM to generate tool plans of various topological structures, the proposed TPL comprises the following key designs. 

\begin{figure}[t]
    \centering
    \includegraphics[width=0.48\textwidth]{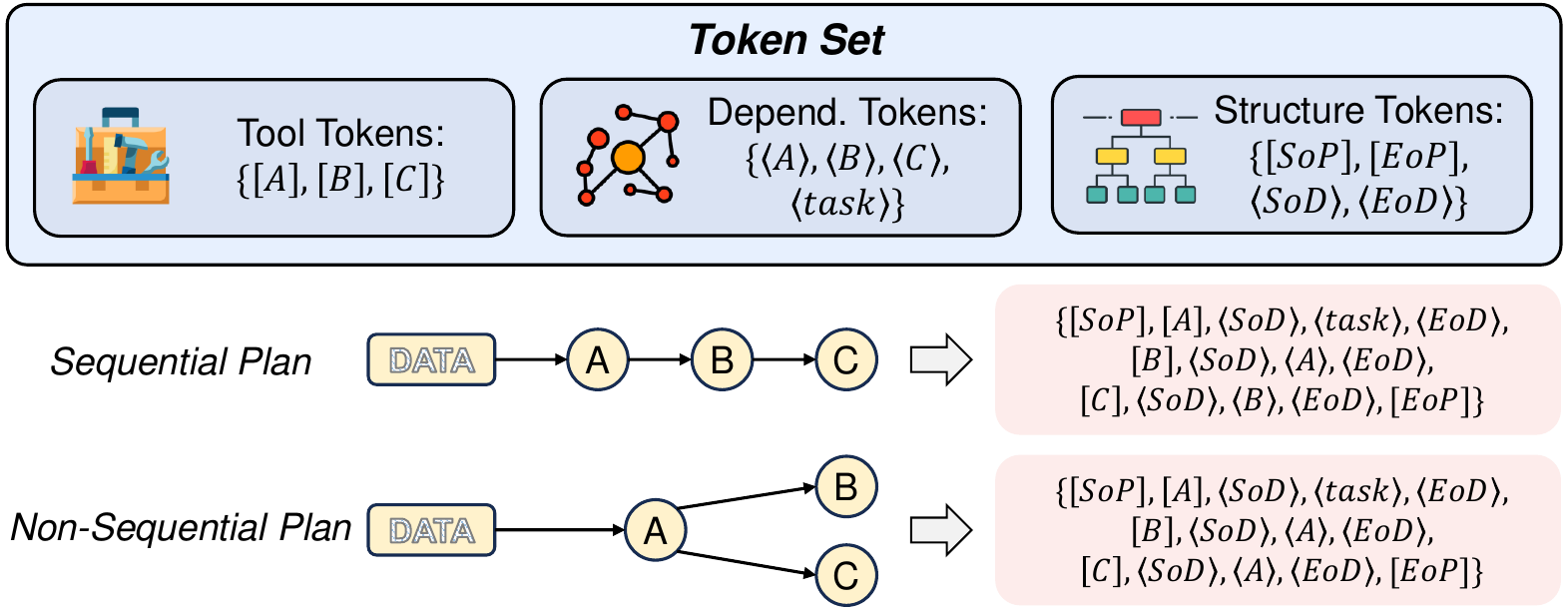}
    \vspace{-0.6cm}
    \caption{Examples of using TPL to encode various tool plans. For illustration, we assume there are three tools $\{A, B, C\}$ and we only consider resource dependencies, where $\langle X\rangle$ means accepting the data provided by $X$ as inputs.}
    \vspace{-0.4cm}
    \label{fig:tpl_demo}
\end{figure}

\noindent \textbf{Anything as token.} Inspired by the work in~\cite{hao2024toolkengpt}, we formulate tools and dependencies as tokens\footnote{The work in~\cite{hao2024toolkengpt} also represents tools as tokens. Our proposed TPL distinguishes itself  largely from~\cite{hao2024toolkengpt} by representing tool dependencies as tokens and explicitly formulating a tool plan as a sequence of tool and dependency tokens (explained later in this section).}, denoted as $[\mathcal{T}]=\{[t_1], [t_2], \cdots\}$ and $\langle \mathcal{D}\rangle = \{\langle d_1\rangle, \langle d_2\rangle, \cdots\}$, respectively. 
Similar to regular word tokens, each tool or dependency token is parameterized by an embedding vector that can be trained to learn the optimal representations for the LLM to understand its semantic information. 
The key benefit of this approach is that it can accommodate new tools or dependencies simply by adding new tokens, allowing the LLM to support massive tools and dependencies.

\noindent \textbf{Plan as sequence.} 
Due to the complexity of DAG structures,  it can be challenging and error-prone for the LLM to directly generate tool plans in the DAG form. 
To tackle this issue, we explicitly transform a DAG tool plan into a sequence of tool and dependency tokens to cater to the sequence prediction nature of the LLM:
\vspace{-0.5cm}

\begin{equation}
\begin{aligned}
    \setlength{\abovedisplayskip}{4pt}
    p&=\{[SoP], \cdots, [t_i],  \langle SoD\rangle, \langle d_1^i\rangle, \cdots, \langle d_j^i\rangle,  \\
     & \;\;\;\;\;\langle EoD\rangle,\cdots, [EoP]\},
    \setlength{\belowdisplayskip}{4pt}
\end{aligned}
\end{equation}
\vspace{-0.2cm}


\noindent where $[t_i]\in[\mathcal{T}]$ denotes the $i$-th tool token, and $\langle d_j^i\rangle\in\langle\mathcal{D}\rangle$ represents the dependency between the $i$-th tool and the previous tool. In our practical implementation, we introduce an additional token $\langle task \rangle$ to indicate a special dependency that a tool accepts the data provided by the task (\eg, an image) as inputs. 
In particular,  $[SoP]$, $[EoP]$ and $\langle SoD\rangle$, $\langle EoP \rangle$ are special structure tokens indicating the start, end of the plan and start, end of the dependencies associated with a tool, respectively. These structure tokens are used to facilitate the LLM to effectively perceive the plan structural information.
Through such a flexible representation, our TPL can represent tool plans of diverse structures, enhancing the non-sequential planning ability of the LLM. Figure~\ref{fig:tpl_demo} exemplifies the use of TPL for encoding various tool plans.

\subsection{Cost-Aware Offline Reinforcement Learning}
\label{subsec:caorl}
With the proposed TPL, the LLM is capable to devise various tool plans simply by  predicting tool and dependency tokens. 
To optimize the plan generation process, CATP-LLM then fine-tunes the LLM through the efficient CAORL mechanism. 
Specifically, CAORL includes: i) a context augmentation scheme to effectively integrate tool cost information into the LLM’s input context; ii) an offline RL-based fine-tuning algorithm that guides the LLM to achieve the optimal performance-cost trade-off in tool planning.

\vspace{-0.2cm}
\subsubsection{Context Augmentation}
\label{subsubsec:context_augmentation}

\begin{figure}[t]
    \centering
    \includegraphics[width=0.48\textwidth]{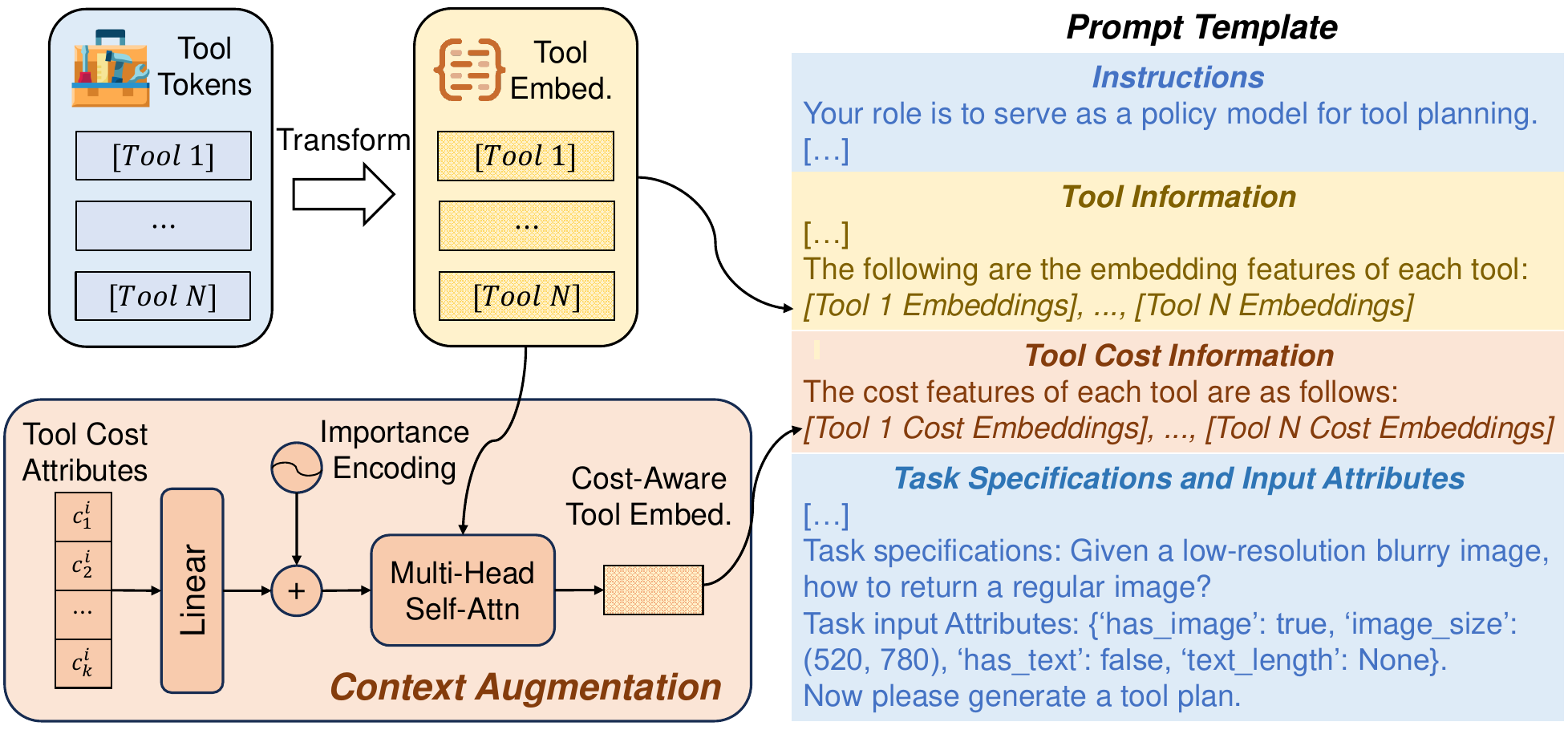}
    \vspace{-0.6cm}
    \caption{Illustrations of the prompt template and context augmentation design. Complete prompts are displayed in Appendix $\S$\ref{appendix:caorl_details}.}
    \vspace{-0.6cm}
    \label{fig:context_augment}
\end{figure}

Figure~\ref{fig:context_augment} illustrates the prompt template designed to provide the LLM with rich information for effective planning. We  integrate the embedding features of each tool into the prompt for the LLM    to understand their functionalities. Moreover, it is also essential to incorporate tool cost information into the prompt for effective cost-aware planning. However, since tool costs can vary with different sizes of task inputs, merely writing the cost information in the prompt fails to establish the connection between input data sizes and tool costs. To address this issue, we further propose a context augmentation scheme that effectively transforms the tool cost information into a set of cost-aware embedding features, as shown in Figure~\ref{fig:context_augment}. 


The details of the context augmentation scheme are elaborated as follows. First, we categorize the input sizes into $\{1, \cdots, k\}$ levels as two very close sizes will not make a big difference (\eg, resolution 480$\times$520 \vs 490$\times $510, batch size 32 \vs 36). Subsequently, each tool is assigned with a cost attribute vector $c(t_i)= (c_1^i, \cdots, c_k^i)$, indicating its execution costs at different levels of input sizes. Afterward, we employ a linear layer to extract features from the tool cost attributes. We then encode the importance of cost features by adding the following importance vector:
\begin{equation}
    \setlength{\abovedisplayskip}{3pt}
    v_{i- 1}=\cos\frac{\pi(i-l)}{2k}, i\in\{1,\cdots,k\},
    \setlength{\belowdisplayskip}{4pt}
\end{equation}
where $l$ denotes the level of the current task inputs. 
This importance vector allows the importance of cost features to gradually increase based on their distance to the current size level. 
Finally, a multi-head self-attention module is utilized for feature fusion, producing a set of cost-aware features.


The remaining problem here is how to acquire the cost attributes of each tool $c(t_i)$. To tackle this issue, we manually construct various tool plans and ensure that each tool is used at least once in these plans. We then simulate diverse input data at varying size levels, feed the simulated data to each plan, and record the execution costs of each tool (\eg, execution time and memory consumption), which ultimately yields the cost attributes of each tool.


\subsubsection{Learning Through Offline RL}
\begin{figure}[t]
    \centering
    \includegraphics[width=0.45\textwidth]{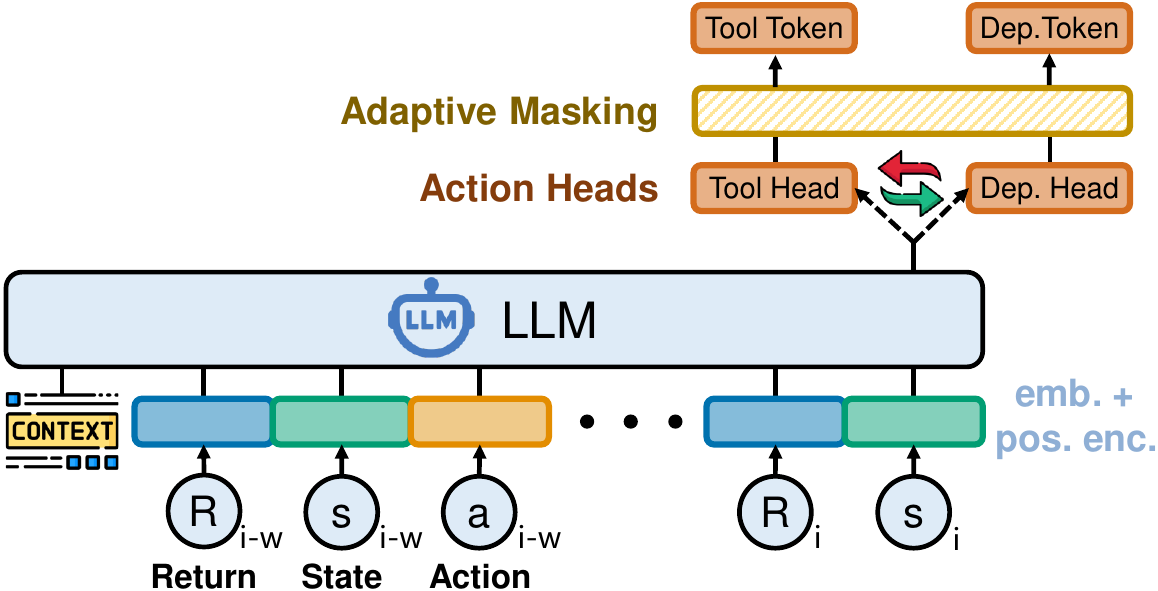}
    \vspace{-0.2cm}
    \caption{Illustrations of the decision transformer~\cite{chen2021decision} based learning algorithm.}
    \vspace{-0.4cm}
    \label{fig:dt_framework}
\end{figure}

Despite augmenting the LLM with rich input context, achieving cost-aware tool planning remains challenging due to the inherent trade-off between plan performance and costs. 
Hence, it is imperative to fine-tune the LLM to acquire the domain knowledge for cost-aware planning. 
Notably, we observe that the selection of a tool can have cascading effects on the subsequent selection. 
For instance, selecting a powerful yet costly tool may lead to avoiding the selection of expensive tools later to prevent costs from becoming unaffordable.
This suggests that the cost-aware tool planning is essentially a sequential decision problem, a domain where reinforcement learning (RL)~\cite{wang2022deep} excels. 
Thus, we design the fine-tuning algorithm based on RL, as illustrated in Figure~\ref{fig:dt_framework} and detailed below.

\noindent \textbf{State.} 
At timestep $i$, the LLM will receive a state $s_i$ defined as a token sequence of the current tool  plan:
\begin{equation}
    \setlength{\abovedisplayskip}{3pt}
    s_{i}=\{tk_1, \cdots, tk_i\},
    \setlength{\belowdisplayskip}{3pt}
\end{equation}
where $tk_i$ denotes a tool, dependency or structure token.


\noindent \textbf{Action.} Based on state $s_i$ and input context, the LLM takes an action $a_i$ representing a tool token or dependency token. We design two output heads for the LLM to generate actions: \textit{tool head} for tool tokens and \textit{dependency head} for dependency tokens. In particular, we extend the output space of tool head and dependency head with $[EoP]$ and $\langle EoD\rangle$ tokens, respectively. During plan generation, if the last token in state $s_i$ is $[SoP]$ or $\langle EoD\rangle$, the LLM will use the tool head to predict the next tool token. This token will be added to the plan along with the $\langle SoD\rangle$ token, which forms a new state. Afterward, the LLM will switch to the dependency head to predict dependency tokens. It will switch back to the tool prediction head until the $\langle EoD\rangle$ token is predicted. The plan generation process will terminate as long as the LLM predicts the $[EoP]$ token with the tool head.

\noindent\textbf{Adaptive masking.} To improve the plan generation efficiency, we further introduce an adaptive masking scheme to dynamically filter out invalid tool or dependency tokens based on a set of heuristic rules. For instance, a tool cannot only establish a dependency with a non-existent tool in the current plan. 
While conceptually simple, this approach reduces the action space for the LLM and thereby improves its plan generation efficiency.

\noindent \textbf{Reward with intermediate feedback.} 
Upon taking action $a_i$, the LLM will receive a reward $r_i$ formally defined as:
\begin{equation}
\label{eq:reward}
    \setlength{\abovedisplayskip}{3pt}
    r_i=\left\{
\begin{aligned}
-(1-\alpha)C(p_i) & , & \text{if } a_i\neq[EOP], \\
\alpha  P(p_i) - (1-\alpha) C(p_i) & , & \text{if } a_i=[EOP],
\end{aligned}
\right.
    \setlength{\belowdisplayskip}{3pt}
\end{equation}
where $p_i = s_i \cup \{a_i\}$ denotes the plan updated with action $a_i$, $P(p_i)$ calculates the performance scores of $p_i$ on the target task (\eg, BERT scores~\cite{zhang2019bertscore} for image captioning), $C(p_i)$ measures the execution costs of $p_i$ (\eg, execution time), and $\alpha\in (0, 1)$ serves as a weight parameter to balance the importance of two metrics ($\alpha=0.5$ by default). 
In contrast to prior studies~\cite{ge2024openagi,qiao2023making} where the LLM only receives feedback upon completing a plan, we allow the execution of an incomplete plan during the fine-tuning stage to collect its execution costs\footnote{Note that unlike execution costs, obtaining performance scores for an incomplete plan may be impractical. For instance, for the task in Figure~\ref{fig:example_figure}, an incomplete plan like ``colorization$\rightarrow$super\_resolution'' cannot address the task due to the output type mismatch (image \vs text). } as the feedback replayed to the LLM. This guides the LLM toward minimizing execution costs while maximizing task scores in plan generation.

\noindent\textbf{Learning algorithm.}  
We fine-tune the LLM for plan generation based on the offline RL algorithm decision transformer (DT)~\cite{chen2021decision}  because of its efficiency.
DT reformulates RL as a sequence modeling problem, which seamlessly aligns with the sequence modeling nature of LLM and the proposed TPL. 
As illustrated in Figure~\ref{fig:dt_framework}, in this algorithm, the LLM models the distribution of actions conditioned on the specific states and returns, where return $R_i=\sum_{j=i} r_j$ represents the cumulative rewards expected to receive from state $s_i$. After training, the LLM can generate  actions to achieve the desired return~\cite{chen2021decision,wu2024netllm}. More details about the algorithm can be found in Appendix $\S$\ref{appendix:caorl_details}.

In lack of public tool plan datasets for fine-tuning, we propose a dataset generation pipeline to collect the necessary training data. 
We begin by constructing a graph where each node represents a tool and each edge denotes the dependency between two tools. For each task, we then generate a tool plan by searching for a sub-graph within the tool graph that could potentially solve the target task. To reduce the search space, we prune tools that are not relevant to the target task (\eg, image captioning tool for an object detection task). Besides, since not all generated plans are valid (\ie, executable), we will filter out those invalid plans and utilize the rest as the plan dataset for fine-tuning.

\section{OpenCATP Dataset}
\label{sec:benchmark}

\ICCVRevision{In this section, we elaborate the details of the proposed OpenCATP dataset.}

\noindent\textbf{Evaluation tasks.} \ICCVRevision{OpenCATP features a range of compositional tasks for the comprehensive evaluation of cost-aware tool planning. 
It includes 87 sequential tasks that can be solved with sequential plans, which are constructed from the popular sequential planning dataset OpenAGI~\cite{ge2024openagi}. These tasks can be categorized as: \textit{image-in-image-out} tasks (\eg, transforming a blurry image to a regular image) and \textit{image-in-text-out} tasks (\eg, image classification and captioning). 
More importantly, it introduces 24 challenging non-sequential tasks, which are absent in previous datasets. These tasks can be classified into two categories: \textit{image-in-text-out} tasks and \textit{image-in-image-text-out} tasks (\eg, regularizing a noisy image and identifying the object in it). The detailed construction of sequential/non-sequential tasks is provided in Appendix $\S$\ref{appendix:opencatp}, with Table~\ref{tab:example_opencatp} offering examples of different types of tasks.}

\ICCVRevision{Note that each task in OpenCATP is composed of 100 samples of varying input sizes, resulting in 8,700 and 2,400 data points for the evaluation of sequential and non-sequential planning, respectively. Besides, OpenCATP includes 10 open-sourced models as tools (e.g. image captioning model and deblurring model), offering diverse options for tool composition for the LLM to schedule to solve different tasks, as detailed in Appendix $\S$\ref{appendix:opencatp}.}

\noindent\textbf{Evaluation metrics.} Another key feature of OpenCATP is that it defines a cost model to comprehensively reflect the costs of tool plans, inspired by the Function as a Service (FaaS)~\cite{shahrad2019architectural,awslambda} platforms. 
To be specific, FaaS providers have developed a mature pricing model to charge for the service of running a function based on the execution time and resource consumption. 
Hence, we take cues from the FaaS platforms and design a pricing model in OpenCATP to quantify the overall costs of a tool plan, which jointly considers execution time and instant/constant memory consumption (detailed in Appendix $\S$\ref{appendix:opencatp}). 
Based on this model, we introduce a unified metric Quality of Plan (QoP) in OpenCATP to quantify the quality of a tool plan:
\begin{equation}
\label{eq:qop}
    \setlength{\abovedisplayskip}{3pt}
QoP=\alpha P_{task}(p)-(1-\alpha)C_{price}(p),
    \setlength{\belowdisplayskip}{3pt}
\end{equation}
\ICCVRevision{where $P_{task}(p)$ quantifies the performance of a executed plan on the target task, which can be ViT scores~\cite{ge2024openagi} for image-out tasks and BERT scores~\cite{zhang2019bertscore} for text-out tasks; $C_{price}(p)$ calculates the execution prices of a tool plan reflecting its overall costs; $\alpha \in (0, 1)$ is the weight parameter.}
We employ  min-max normalization on these two metrics when calculating QoP to make sure their values fall in the same scale. Notably, the min/max values are derived from empirical measurement.

\definecolor{rcolor}{RGB}{255,224,224}
\definecolor{ccolor}{RGB}{234,235,255}
\newcommand{\cellwidth}{1.15cm}
\newcommand{\taskcellwidth}{1.15cm}
\newcommand{\qopcellwidth}{1.1cm}
\begin{table*}[t]
\setlength{\tabcolsep}{1pt}
\centering
\scriptsize
\begin{tabular}{ c  |
m{\taskcellwidth}<{\centering}  m{\cellwidth}<{\centering} m{\cellwidth}<{\centering}>{\columncolor{ccolor}}m{\qopcellwidth}<{\centering}| 
m{\taskcellwidth}<{\centering} m{\cellwidth}<{\centering}  m{\cellwidth}<{\centering}  >{\columncolor{ccolor}}m{\qopcellwidth}<{\centering}  |
m{\taskcellwidth}<{\centering}  m{\cellwidth}<{\centering}  m{\cellwidth}<{\centering} >{\columncolor{ccolor}}m{\qopcellwidth}<{\centering}}
\toprule

     & \multicolumn{4}{c|}{\textbf{Image-in-Image-out Tasks}}             & \multicolumn{4}{c|}{\textbf{Image-in-Text-out Tasks}}                                                                                  & \multicolumn{4}{c}{\textbf{Average}}  \\
                             \textbf{Methods} & Task Scores $\uparrow$          & Exec. Prices (\textdollar) $\downarrow$ & Exec. \; Time (s) $\downarrow$ & QoP $\uparrow$& Task Scores $\uparrow$          & Exec. Prices (\textdollar) $\downarrow$ & Exec. \; \; Time (s) $\downarrow$ & QoP $\uparrow$& Task Scores  $\uparrow$         & Exec. Prices (\textdollar) $\downarrow$ & Exec. \;  Time $\downarrow$ (s) & QoP $\uparrow$\\
                             \midrule

Zero-Shot (GPT-3.5) & 0.694 & 0.170 & 2.457 & 0.197 & 0.585 & 0.134 & 2.526 & 0.174 & 0.639 & 0.152 & 2.491 & 0.186\\
Few-shot (GPT-3.5) &  0.715 & 0.148 & 2.073 & 0.227 & 0.618 & 0.123 & 2.331 & 0.201 & 0.666 & 0.136 & 2.202 & 0.214 \\
HuggingGPT (GPT-3.5) & 0.674 & 0.127 & 1.782 & 0.226 & 0.617 & 0.087 & 1.656 & 0.231 & 0.645 & 0.107 & 1.714 & 0.228\\
HYDRA (GPT-3.5) & 0.558 & 0.035 & 0.460 & 0.248 & 0.625 & 0.062 & 1.144 & 0.258 & 0.591 & 0.049 & 0.802 & 0.253 \\
Zero-Shot (GPT-4) & 0.699 & 0.163 & 2.260 & 0.206 & 0.616 & 0.150 & 2.844 & 0.176 & 0.658 & 0.156 & 2.552 & 0.190 \\
Few-shot (GPT-4) & 0.714 & 0.148 & 2.098 & 0.225 & 0.616 & 0.133 & 2.509 & 0.190 & 0.665 & 0.141 & 2.304 & 0.208 \\
HuggingGPT (GPT-4) & 0.675 & 0.131 & 1.893 & 0.221 & 0.616 & 0.085 & 1.615 & 0.232 & 0.646 & 0.108 & 1.754 & 0.227 \\
HYDRA (GPT-4) & 0.597 & 0.050 & 0.615 & 0.253 & 0.623 & 0.058 & 1.079 & 0.259 & 0.610 & 0.054 & 0.847 & 0.256 \\
IFT (Llama2-7B) & 0.616 & 0.245 & 4.441 & 0.092 & 0.651 & 0.176 & 2.619 & 0.170 & 0.634 & 0.210 & 3.530 & 0.131 \\
RLTF (Llama2-7B) & 0.743 & 0.234 & 4.310 & 0.165 & 0.687 & 0.166 & 2.284 & 0.196 & 0.715 & 0.200 & 3.297 & 0.181 \\
\rowcolor{rcolor}
{\textbf{CATP-LLM}} (Llama2-7B) & 0.711 & 0.106 & 1.866 & {\textbf{0.262}}  & 0.651 & 0.074 & 0.876 & {\textbf{0.260}}   & 0.681 & 0.090 & 1.371 & {\textbf{0.261}}      \\
\bottomrule
\end{tabular}
\vspace{-0.3cm}
\caption{Compare CATP-LLM with other methods on different sequential planning tasks. Arrow $\uparrow$ / $\downarrow$ means higher/lower is better. }
\label{tab:main_results_seq}
\vspace{-0.3cm}
\end{table*}

\begin{table*}[t]
\centering
\scriptsize
\setlength{\tabcolsep}{1pt}
\begin{tabular}{ c  |
m{\taskcellwidth}<{\centering}  m{\cellwidth}<{\centering} m{\cellwidth}<{\centering}>{\columncolor{ccolor}}m{\qopcellwidth}<{\centering}| 
m{\taskcellwidth}<{\centering} m{\cellwidth}<{\centering}  m{\cellwidth}<{\centering} >{\columncolor{ccolor}} m{\qopcellwidth}<{\centering}  |
m{\taskcellwidth}<{\centering}  m{\cellwidth}<{\centering}  m{\cellwidth}<{\centering} >{\columncolor{ccolor}}m{\qopcellwidth}<{\centering}}
\toprule
    & \multicolumn{4}{c|}{\textbf{Image-in-Text-out Tasks}}             & \multicolumn{4}{c|}{\textbf{Image-in-Image-Text-out Tasks}}                                                                                  & \multicolumn{4}{c}{\textbf{Average}}                                                                                         \\
                             \textbf{Methods} & Task Scores $\uparrow$          & Exec. Prices (\textdollar) $\downarrow$ & Exec. \;Time (s) $\downarrow$ & QoP $\uparrow$& Task Scores $\uparrow$          & Exec. Prices (\textdollar) $\downarrow$ & Exec.\; Time (s) $\downarrow$ & QoP $\uparrow$& Task Scores  $\uparrow$         & Exec. Prices (\textdollar) $\downarrow$ & Exec.\; Time $\downarrow$ (s) & QoP $\uparrow$\\
                             \midrule

Zero-Shot (GPT-3.5) & 0.506 & 0.178 & 2.632 & 0.095 & 0.150 & 0.121 & 1.938 & -0.031 & 0.328 & 0.149 & 2.285 & 0.032\\
Few-shot (GPT-3.5) & 0.318 & 0.091 & 1.686 & 0.079 & 0.244 & 0.133 & 1.991 & 0.004 & 0.281 & 0.112 & 1.839 & 0.041 \\
HuggingGPT (GPT-3.5) & 0.349 & 0.164 & 2.108 & 0.029 & 0.267 & 0.086 & 1.049 & 0.057 & 0.308 & 0.125 & 1.579 & 0.043\\
HYDRA (GPT-3.5) & 0.184 & 0.027 & 0.210 & 0.068 & 0.269 & 0.087 & 1.387 & 0.058 & 0.227 & 0.057 & 0.798 & 0.064\\
Zero-Shot (GPT-4) & 0.654 & 0.252 & 3.619 & 0.105 & 0.315 & 0.206 & 3.169 & -0.024 & 0.485 & 0.229 & 3.394 & 0.040\\
Few-shot (GPT-4) & 0.635 & 0.197 & 2.530 & 0.143 & 0.330 & 0.142 & 1.991 & 0.039 & 0.482 & 0.170 & 2.261 & 0.091\\
HuggingGPT (GPT-4) & 0.624 & 0.199 & 2.670 & 0.136 & 0.326 & 0.138 & 1.938 & 0.040 & 0.475 & 0.168 & 2.304 & 0.088\\
HYDRA (GPT-4) & 0.326 & 0.038 & 0.712 & 0.129 & 0.347 & 0.118 & 2.005 & 0.069 & 0.337 & 0.077 & 1.358 & 0.099 \\
IFT (Llama2-7B) & 0.176 & 0.158 & 1.485 & -0.051 & 0.516 & 0.215 & 2.676 & 0.068 & 0.346 & 0.186 & 2.080 & 0.008\\
RLTF (Llama2-7B) & 0.323 & 0.150 & 1.400 & 0.029 & 0.478 & 0.226 & 2.553 & 0.039 & 0.401 & 0.188 & 1.976 & 0.034\\
\rowcolor{rcolor}
{\textbf{CATP-LLM}} (Llama2-7B) & 0.641 & 0.093 & 0.753 & {\textbf{0.238}}  & 0.528 & 0.134 & 0.706 & {\textbf{0.145}}   &0.585 & 0.114 & 0.730 & {\textbf{0.192}}      \\
\bottomrule
\end{tabular}
\vspace{-0.3cm}
\caption{Compare CATP-LLM with other methods on different non-sequential planning tasks. Arrow $\uparrow$ / $\downarrow$ means higher/lower is better. }
\label{tab:main_results_nonseq}
\vspace{-0.5cm}
\end{table*}

\section{Experiments}
\label{sec:experiments}
\subsection{Setup}
\label{subsec:setup}

\noindent\textbf{Datasets.} 
\ICCVRevision{For sequential planning, we adopt the sequential tasks of OpenCATP, which are originally constructed from the popular sequential planning dataset OpenAGI~\cite{ge2024openagi}. For non-sequential planning, we utilize the tasks developed from OpenCATP. }

\noindent\textbf{Metrics.} Leveraging the OpenCATP evaluation scheme, we use Quality of Plan (QoP), task performance scores, and execution prices defined in Eq.(\ref{eq:qop}) as the evaluation metrics. 
By default, we set the weight parameter in Eq.(\ref{eq:qop}) $\alpha = 0.5$.
Moreover, we also introduce execution time as the metric to intuitively reflect the overhead of tool plans.

\noindent\textbf{Implementation.} When implementing our CATP-LLM framework over OpenCATP, we use the QoP function in Eq.(\ref{eq:qop}) for the reward calculation. That said, the performance scores and costs of the reward function in Eq.(\ref{eq:reward}) are consistent with those of the QoP function in Eq.(\ref{eq:qop}). 
More details can be found in Appendix $\S$\ref{appendix:exp_impl}.

\noindent\textbf{Baselines.} We compare  CATP-LLM with other representative frameworks. Specifically, for prompt engineering based paradigm, we implement the following baselines for comparison: zero-shot learning (Zero-Shot)~\cite{kojima2022large}, few-shot learning (Few-Shot)~\cite{lu2024chameleon,suris2023vipergpt}, HuggingGPT~\cite{shen2024hugginggpt} and HYDRA~\cite{ke2024hydra}. HuggingGPT is similar to few-shot learning but involves a dedicated planning procedure. It first provides in-context examples to prompt the LLM to decompose a task into sub-tasks in JSON format and next associates each sub-task with a tool for task solving. \ICCVRevision{HYDRA prompts the LLM to generate codes for scheduling tools, where each tool is represented as a python API.} For the fine-tuning paradigm, the following methods are implemented for comparison: instruction fine-tuning (IFT)~\cite{hao2024toolkengpt,schick2024toolformer,yang2024gpt4tools} and RLTF~\cite{ge2024openagi}. IFT refines the LLM's behavior by fine-tuning it over labeled data containing task descriptions and tool plans. RLTF employs an online RL algorithm to fine-tune the LLM, where the LLM learns to optimize its decisions based on the feedback of final plan execution. 
We use QoP in Eq.(\ref{eq:qop}) as the feedback signal for RLTF.

For the prompt-based methods, we use GPT-3.5~\cite{chatgpt} and GPT-4~\cite{achiam2023gpt} as the backbone LLMs. To ensure fairness, we also provide tool cost information in the prompts for these methods, as detailed in Appendix $\S$\ref{appendix:exp_impl}. For the fine-tuning methods, including CATP-LLM, the open-sourced Llama2-7B~\cite{touvron2023llama} is employed as the default LLM. Besides, we use LoRA~\cite{hu2021lora} with rank 64 to fine-tune Llama2-7B for all fine-tuning methods. Furthermore, we adapt our tool planning language for Zero-Shot, Few-Shot, IFT and RLTF methods, since we find that they achieve poor performance in  non-sequential planning without an appropriate formatted language to support the generation of non-sequential plans. 

\noindent\textbf{Hardware.} We conduct all experiments on an Intel Xeon Sliver 20C40T server with two 32GB V100 GPUs. 

\subsection{Main Results}
\label{subsec:main_results}
\noindent\textbf{Sequential Planning.} Table~\ref{tab:main_results_seq} compares CATP-LLM with other methods for sequential planning. 
We can see that CATP-LLM consistently maintains low execution costs of tool plans while achieving comparable performance across all cases. Thus, CATP-LLM outperforms GPT-based methods even when using Llama2-7B as its backbone, with 1.5\%-40.6\% improvements in QoP on average. Additionally, CATP-LLM also achieves 99.0\% and 44.3\% higher average QoP than IFT and RLTF, respectively. On one hand, IFT fine-tunes the LLM on labeled data of task-plan pairs, which faces difficulty in generalizing to new tasks. On the other hand, unlike CATP-LLM that enables the LLM to evaluate plan costs each time a decision is made, RLTF only allows the LLM to access the plan quality when the plan is completely generated and executed. Consequently, this can mislead the LLM to aggressively optimize plan performance with the price of high execution costs.

\begin{table}[t]
    \centering
    \scriptsize
    \setlength{\tabcolsep}{0pt}
    \begin{tabular}{m{1.7cm}<{\centering} |  m{1.66cm}<{\centering}  
     m{1.66cm}<{\centering}   m{1.73cm}<{\centering} >{\columncolor{ccolor}} m{1.56cm}<{\centering}}
    \toprule
        Method & Task Scores & Exec. Prices (\textdollar) & \% of Valid Plans & QoP  \\ 
        \midrule
        w/o context augmentation & 0.654 & 0.092  & 100\% & 0.246 \\ 
        w/o intermediate feedback & 0.708 & 0.164 & 100\% & 0.209  \\ 
        w/o adaptive masking & 0.634 & 0.086 & 70.2\% & 0.242 \\ 
        \rowcolor{rcolor}
        {\textbf{CATP-LLM}}  & 0.681 & 0.090  & {\textbf{100\%}} & {\textbf{0.261}}\\ 
        \bottomrule
    \end{tabular}
    \vspace{-0.3cm}
    \caption{Explore the importance of different techniques in CAORL algorithm of CATP-LLM.}
    \label{tab:ablation_components}
    \vspace{-0.54cm}
\end{table}

\noindent\textbf{Non-sequential Planning.} 
As shown in Table~\ref{tab:main_results_nonseq}, CATP-LLM demonstrates significantly better performance than other methods in non-sequential planning across all tasks, with 92.0\%-375.4\% higher QoP on average.
\ICCVRevision{Moreover, we empirically observe that GPT-3.5 generates many invalid plans due to the complexity of non-sequential tasks: 41.7\% for Zero-Shot, 75\% for Few-Shot, and 66.7\% for HuggingGPT. For example, it may link an image-input tool behind a text-output tool.}
In comparison, CATP-LLM succeeds to guarantee the validity of 100\% tool plans even with the 7B Llama2 as the backbone. 
It is worth noting that while some methods (\eg, Zero-Shot and HYDRA) may achieve comparable execution prices to CATP-LLM, their execution time is much higher. 
This is because most of their plans generated maintain a sequential structure, and thus cannot be executed in parallel to reduce  execution time.

\noindent\textbf{Qualitative Results.} Figure~\ref{fig:example_figure} illustrates the cost-aware tool planning capability of CATP-LLM. For a complex task requiring multiple outputs (\ie an image caption in German and an object name), CATP-LLM accurately devises a non-sequential plan that can be executed in parallel to reduce execution time. In contrast, the plans generated by GPT-3.5 follow  a sequential structure or lack the necessary machine translation tool, resulting in task failure and low performance scores. While HuggingGPT with GPT-4 includes a costly super-resolution tool, CATP-LLM determines to omit it, as this tool will not significantly enhance performance given that the image size is not very small, but will drastically increase costs. Thus, CATP-LLM efficiently reduces plan costs without sacrificing performance, achieving higher QoP than HuggingGPT.

\subsection{Ablation Study}
\label{subsec:ablation}

\noindent\textbf{Importance of different techniques in CAORL.} The proposed fine-tuning algorithm CAORL plays an significant role in enabling CATP-LLM for cost-aware tool planning. Hence, we explore the contributions of various important techniques in CAORL to CATP-LLM, including context augmentation, intermediate feedback signals, and adaptive masking. As shown in Table~\ref{tab:ablation_components}, the context augmentation and intermediate feedback are essential for reducing execution costs, while adaptive masking ensures plan validity.

\begin{figure}
    \centering
    \includegraphics[height=0.13\textheight]{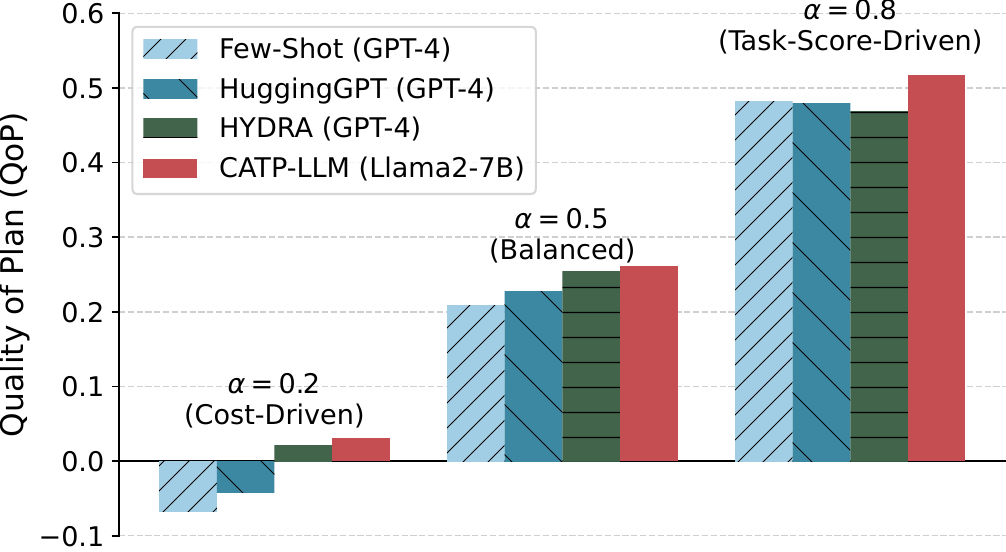}
    \vspace{-0.3cm}
    \caption{Compare CATP-LLM with representative baselines across various values of $\alpha$.}
    \label{fig:ablation_alpha}
    \vspace{-0.4cm}
\end{figure}

\noindent\textbf{Generalization to various QoP definitions.} By default, we set $\alpha$ in QoP to 0.5, indicating equal importance for plan performance and costs. In fact, CATP-LLM is a general framework that can adapt to different QoP definitions based on practical applications. We verify its generalizability by varying $\alpha$ in QoP and comparing its performance to GPT-4-powered methods. As shown in Figure~\ref{fig:ablation_alpha}, CATP-LLM consistently yields the highest QoP scores across all cases.

\begin{figure}
    \centering
    \includegraphics[height=0.135\textheight]{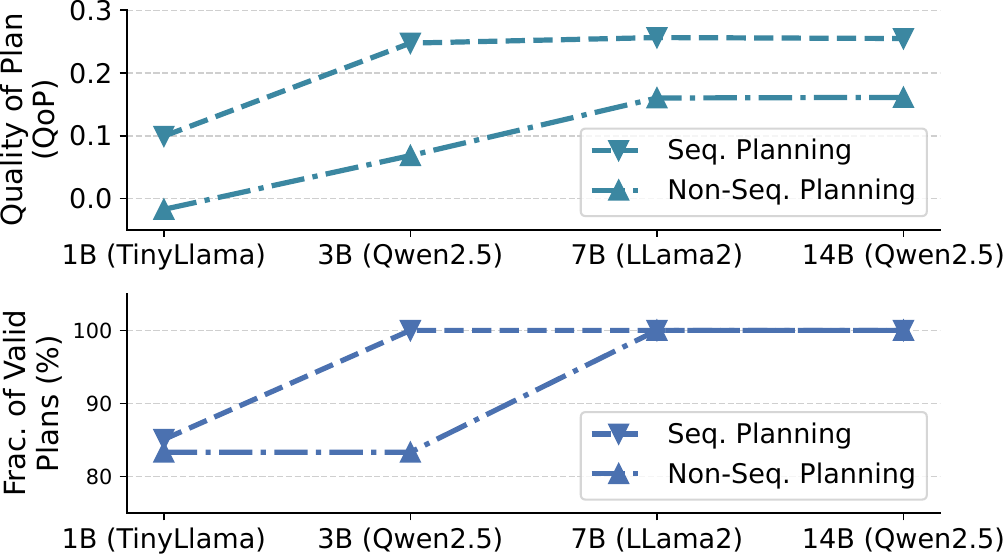}
    \vspace{-0.3cm}
    \caption{Explore the impacts of LLM sizes on  CATP-LLM.}
    \label{fig:ablation_size}
    \vspace{-0.6cm}
\end{figure}

\noindent\textbf{Impacts of LLM parameter sizes.} Finally, we explore the impacts of LLM sizes on CATP-LLM's efficacy. We use TinyLlama-1B~\cite{zhang2024tinyllama}, Qwen-3B~\cite{hui2024qwen2}, Llama2-7B~\cite{touvron2023llama} and Qwen-14B~\cite{hui2024qwen2} as backbone LLMs. As shown in Figure~\ref{fig:ablation_size}, Qwen-3B demonstrates comparable performance to Llama2-7B in sequential planning, but Llama2-7B significantly outperforms others in complex non-sequential tasks. This suggests that for relatively simple sequential tasks, smaller LLMs are sufficient for planning, while larger LLMs are recommended for more complex tasks. 

\vspace{-0.2cm}
\section{Conclusion}
\vspace{-0.1cm}
\label{sec:conclusion}
In this paper, we propose the first cost-aware tool planning framework CATP-LLM. It features a tool planning language to enhance the LLM to create non-sequential plans of multiple branches for efficient concurrent execution. Building on this language, it further introduces a cost-aware offline reinforcement learning algorithm to fine-tune the LLM to optimize the trade-off between plan performance and costs. 
We then present OpenCATP, the first dataset for cost-aware planning evaluation. Experiments on OpenCATP show that CATP-LLM significantly outperforms GPT series in cost-aware tool planning, even with a 7B LLM as its backbone.


\section*{Acknowledgements}
We thank the anonymous ICCV reviewers for their invaluable feedback. Duo Wu would also like to thank his colleagues in the lab, Shuzhao Xie, Rongwei Lu, Jiajun Luo and Xiaojie Li, for their constructive comments on improving the quality of this paper. This work was supported in part by National Key Research and Development Project of China (Grant No. 2023YFF0905502), National Natural Science Foundation of China (Grant No. 92467204, 62472249 and 62402264), and Shenzhen Science and Technology Program (Grant No. JCYJ20220818101014030 and KJZD20240903102300001). 

{
    \small
    \bibliographystyle{ieeenat_fullname}
    \bibliography{main}
}

\clearpage
\renewcommand{\thesection}{\Alph{section}}
\setcounter{section}{0}
\setcounter{table}{0}   
\setcounter{figure}{0}
\renewcommand{\thetable}{\thesection{}.\arabic{table}}
\renewcommand{\thefigure}{\thesection{}.\arabic{figure}}

\section{Details of CAORL}
\label{appendix:caorl_details}
\noindent \textbf{Prompts of context augmentation.}  The complete prompt of context augmentation scheme is illustrated belows.
\lstset{style = singlecolumnstyle}
\begin{lstlisting}
# Instructions

 Your role is to serve as a policy model for tool planning. Given a task and a set of tools, you need to select some tools and determine the execution plan of tools that can be executed in sequential or parallel to solve the task. Your goal is to generate the tool plans that can optimize the task performance while minimizing the execution costs.

# Tool Information

 Each tool has its own functionality. Executing a tool will incur some execution costs. Besides, the costs of each tool may vary based on the size of inputs. The following are the embedding features of each tool:

 [Tool 1 Embeddings], ..., [Tool N Embeddings]

# Tool Cost Information

 The cost features of each tool are as follows:
 
 [Tool 1 Cost Embeddings], ..., [Tool N Cost Embeddings]

# Task Specifications and Input Attributes

 Next, you will receive information about the task and the attributes of the current inputs.
 Task specifications: Given an low-resolution blurry image, how to return a regular image?
 Task input Attributes: {'has_image': true, 'image_size': (520, 780), 'has_text': false, 'text_length': None}.
 Now please generate a tool plan.

\end{lstlisting}

\noindent \textbf{Learning algorithms.} The details of the decision transformer~\cite{chen2021decision} based learning algorithm in CAORL are described as follows. Formally, the LLM takes the historical return, state, and action sequences to predict the next action:
\begin{equation}
    \setlength{\abovedisplayskip}{3pt}
    LLM(a_i|\mathcal{I};\{R_{i-w},s_{i-w},a_{i-w},\cdots,R_i,s_i\}), \tag{A.1}
    \setlength{\belowdisplayskip}{3pt}
\end{equation}
where $R_i, s_i, a_i$ represent the return, state and action at timestep $i$, respectively. $\mathcal{I}$ denotes the input context described in Figure~\ref{fig:context_augment} and $\S$\ref{subsubsec:context_augmentation}, while parameter $w$ defines the context window to facilitate the LLM to learn the action distribution. The rationale behind this algorithm is to train the LLM to learn the distribution of actions conditioned on specific states and returns, enabling the LLM to generate actions to achieve the desired return after training~\cite{chen2021decision,wu2024netllm}. In particular, during the inference stage, we specify a target return indicating the desired performance to trigger the LLM to generate actions. In practice, we set the target return as the maximum return observed in the training plan dataset.

Note that instead of directly feeding states, actions, and returns into the LLM, we design three separate linear layers to project them into embedding features, followed by layer normalization~\cite{ba2016layer}. Additionally, each embedding vector is added with a learned positional embedding based on its corresponding timestep.


\section{Details of OpenCATP}
\label{appendix:opencatp}

\subsection{Tool Set}
\ICCVRevision{Following OpenAGI~\cite{ge2024openagi}, the tools OpenCATP for task solving consist of 10 open-source domain expert models of different functions. These models include: sentiment analysis \cite{araci2019finbert}, text summarization \cite{lewis2019bart}, machine translation \cite{raffel2020exploring}, image classification \cite{dosovitskiy2020image}, object detection \cite{carion2020end}, image colorization \cite{zhang2017real}, image super-resolution \cite{conde2022swin2sr}, image denoising \cite{zamir2022restormer}, image deblurring \cite{zamir2022restormer} and image captioning \cite{fang2021instances}.}

\vspace{-0.4cm}
\subsection{Evaluation Tasks}
\ICCVRevision{As mentioned in $\S$\ref{sec:benchmark}, OpenCATP includes 87 sequential tasks and 24 non-sequential tasks, each with 100 data samples. The sequential tasks in OpenCATP are constructed from OpenAGI~\cite{ge2024openagi}, one of the most popular LLM sequential tool planning datasets. As for non-sequential tasks, we use the back-instruct method~\cite{shen2023taskbench} for task construction. Specifically, we construct diverse
 non-sequential tasks by following steps: (1) create a tool  graph based on the toolkit and dependencies between tools;  (2) extract a sub-graph representing a non-sequential plan;  (3) prompt the GPT-4 to generate the descriptions of a task  that can be solved by this plan.
Table~\ref{tab:example_opencatp} provides several concrete examples of various types of tasks in OpenCATP.}

\newcommand{\imagecellheight}{0.11}
\newcommand{\iamgecellwidth}{0.25}
\begin{table*}[t]
    \centering
    \small
    \begin{tabular}{m{3.2cm}<{\centering} |m{4.15cm}<{\centering} |m{4.15cm}<{\centering} |m{4.15cm}<{\centering} }
    \toprule
       \textbf{Task}  &  \textbf{Input Sample} & \textbf{Ground Truth}  & \textbf{Example Plan} \\
       \midrule
       \multicolumn{3}{c}{\textit{Sequential Tasks}} \\
       \midrule 
       \textit{Image-in-Image-out Task:}
        Given a noisy blurry grayscale image, how to return the regular image step by step? & \includegraphics[height=\imagecellheight\textheight]{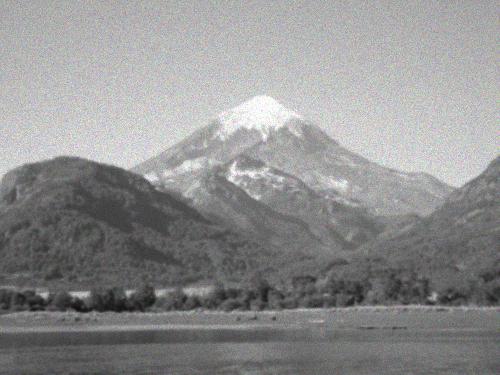} & \includegraphics[height=\imagecellheight\textheight]{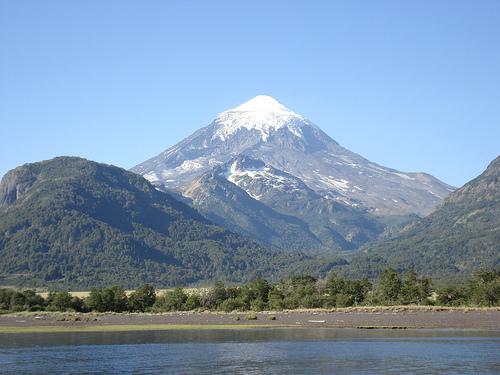} & 
        \includegraphics[width=\iamgecellwidth\textwidth]{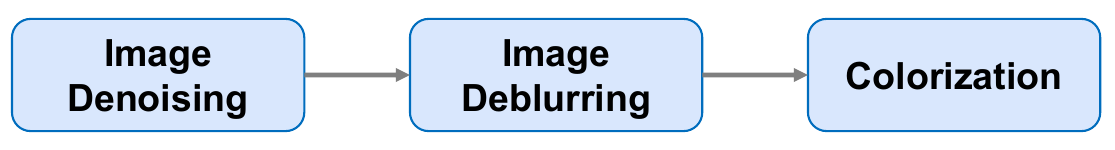}\\
        \midrule
       \textit{Image-in-Text-out Task:}
        Given a noisy blurry grayscale image, how to return the caption in English step by step? & \includegraphics[height=\imagecellheight\textheight]{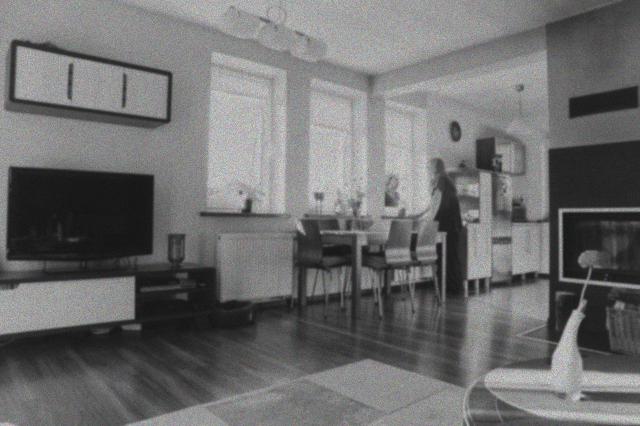} & A woman stands in the dining area at the table. & 
        \includegraphics[width=\iamgecellwidth\textwidth]{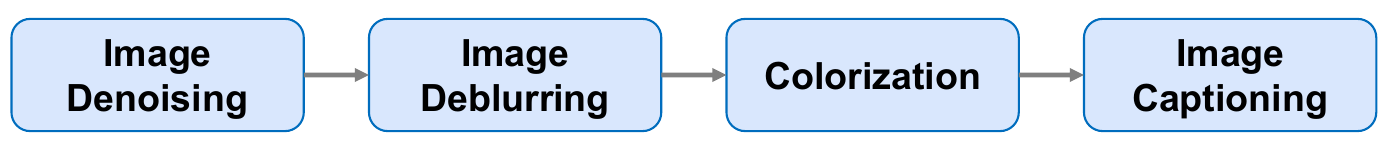}\\
        \midrule
       \multicolumn{3}{c}{\textit{Non-Sequential Tasks}} \\
        \midrule
       \textit{Image-in-Text-out Task:}
        Given a blurry image, how to (1) return the label of the image in English, and (2) return the caption of the image in German step by step? & \includegraphics[height=\imagecellheight\textheight]{figure/opencatp_examples/208-1in.jpg} & (1) Braunbär; (2) Ein kräftiger Grizzly Bär ist im Hintergrund mit Gras zu sehen. & 
        \includegraphics[width=\iamgecellwidth\textwidth]{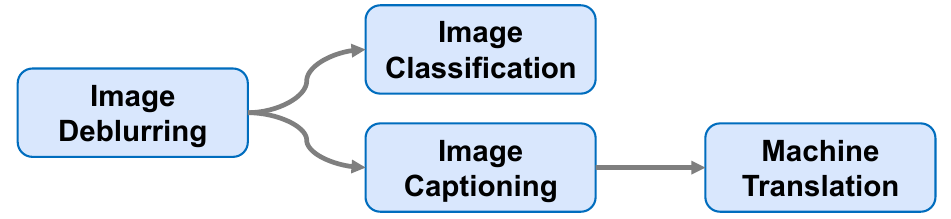}\\
       \midrule
       \textit{Image-in-Image-Text-out Task:}
        Given a blurry noisy image, how to (1) return the regular image, and (2) return the label of the image in German step by step? & \includegraphics[height=\imagecellheight\textheight]{figure/opencatp_examples/208-56in.jpg} & (1) \includegraphics[height=\imagecellheight\textheight]{figure/opencatp_examples/208-56out.jpg}; (2) Schaf. & 
        \includegraphics[width=\iamgecellwidth\textwidth]{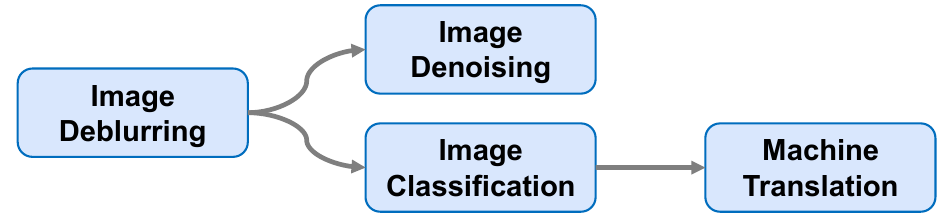}\\
        \bottomrule
    \end{tabular}
    \vspace{-0.3cm}
    \caption{Examples of the sequential and non-sequential tasks and their corresponding data samples in OpenCATP.}
    \label{tab:example_opencatp}
    \vspace{-0.4cm}
\end{table*}

\subsection{Evaluation Metrics}
\label{appendix:evaluate_metrics}
\noindent\textbf{Plan execution prices.} OpenCATP utilizes execution prices to comprehensively reflect the overall costs of tool plans, inspired by the Function as a Service (FaaS) platforms. FaaS is a category of cloud computing that charges consumers in an event-driven manner, where consumers only need to pay for the time and resources used to run their functions, with no prices paid when the functions are idle~\cite{shahrad2019architectural}. This ``pay-as-you-go'' model can result in significant savings, especially for applications with intermittent usage or short-lived tasks, contributing to the market success of FaaS. According to recent statistics~\cite{faasreport}, the global market size of FaaS is estimated to be 17.70  billion USD in 2024 and will reach 44.71   billion USD by 2029.

To accurately charge consumers for running their functions, FaaS platforms like AWS Lambda~\cite{awslambda} have established a mature pricing model based on execution time and resource consumption. The effectiveness of this model has been validated over years in the FaaS industry, which motivates us to adopt execution prices to represent the overall costs of tool plans. Hence, we draw inspiration from the AWS Lambda~\cite{awslambda} and design a pricing model for running a tool as follows:
\begin{equation}
\label{equation:price}
    \setlength{\abovedisplayskip}{3pt}
    \tag{B.1}
    \begin{aligned}
        C_{price}(t)&=price\_per\_run + time\\
        &\times(cpu_{cons}\times price\_cpu_{cons}(t) \\ 
&+cpu_{inst}\times price\_cpu_{inst}\\ 
&+gpu_{cons}\times price\_gpu_{cons}(t)\\\notag
&+gpu_{inst}\times price\_gpu_{inst}).\notag
    \end{aligned}
    \setlength{\belowdisplayskip}{3pt}
\end{equation}
Here, $time$ denotes the execution time in milliseconds of tool $t$. $cpu_{cons}$ ($gpu_{cons}$) denotes the constant CPU (GPU) memory consumption in MB required to load the tool. $cpu_{inst}$ ($gpu_{inst}$) denotes the instant CPU (GPU) memory consumption in MB caused by the computation of the tool. $price\_cpu_{cons}$, $price\_cpu_{inst}$, $price\_gpu_{cons}$, $price\_gpu_{inst}$ calculate the prices in USD associated with the respective memory consumptions. $price\_per\_run$ denotes the prices in USD charged for each execution of the tool. Table~\ref{tab:prices} lists the settings of the pricing parameters\footnote{Note that the existing FaaS platforms do not provide pricing strategy for GPU resources. We then set the GPU prices as three times the CPU prices, according to the article in https://news.rice.edu/news/2021/rice-intel-optimize-ai-training-commodity-hardware}. 

Based on the tool prices, the  execution prices of a tool plan $p$ are defined the sum of prices of each tool in the plan:
\begin{equation}
    \setlength{\abovedisplayskip}{3pt}
    C_{price}(p) = \sum_tC_{price}(t) \tag{B.2}
    \setlength{\belowdisplayskip}{3pt}
\end{equation}

\noindent\textbf{Plan execution time.} For a sequential plan, the execution time is simply the sum of the execution time of each tool in the plan. For example, for plan \ding{172} in Figure~\ref{fig:example_figure}, the execution time is $0.18+3.46+0.13=3.77$(s). For a non-sequential plan that can be executed in parallel, the execution time is defined as the maximum of the sum of tool execution time in each branch of the plan. For example, for plan \ding{175} in Figure~\ref{fig:example_figure}, the execution time is $\max(0.18+0.29+0.16,0.18+0.09)=\max(0.63,0.27)=0.63$ (s).

\noindent\textbf{Task performance of plans.} In OpenCATP, we mainly use ViT scores~\cite{ge2024openagi} and BERT scores~\cite{zhang2019bertscore} to calculate the performance scores of an executed  tool plan on the target task. Specifically, ViT scores are used for tasks that involve image outputs, which measures the cosine similarity between the image generated by the plan and the ground-truth image. BERT scores are used for tasks that involve text outputs, which measures the cosine similarity between the generated texts and ground-truth texts. 

Note that for non-sequential tasks requiring multiple outputs, we calculate the performance scores for each pair of plan outputs and ground-truth outputs, then average these scores to derive the final plan performance scores on the tasks. In particular, if a model produces a sequential plan for a non-sequential task (e.g., Figure~\ref{fig:example_figure}\ding{172}), we assign a 0 score for the missing output pair. For instance, the plan \ding{172} in Figure~\ref{fig:example_figure} will receive a 0 score for the caption output, as it does not generate any captions.

\begin{table}[t]
    \centering
    \footnotesize
    \begin{tabular}{m{2.2cm}<{\centering}  |  m{1.6cm}<{\centering}   m{3.cm}<{\centering}}
    \toprule
        \textbf{Notations} & \multicolumn{2}{c}{\textbf{Values (\textdollar)}} \\ 
        \hline
        $price\_per\_run$ & \multicolumn{2}{c}{2e-7} \\
        \hline
        \multirow{13}{*}{$price\_cpu_{cons}$}& 2.1e-9 & if memory $\le$ 128MB \\
        \cline{2-3}
        & 8.3e-9 & if memory $\le$ 512MB \\
        \cline{2-3}
        & 1.67e-8 & if memory $\le$ 1024MB \\
        \cline{2-3}
        & 2.5e-8 & if memory $\le$ 1536MB \\
        \cline{2-3}
        & 3.33e-8 & if memory $\le$ 2048MB \\
        \cline{2-3}
        & 5e-8 & if memory $\le$ 3072MB \\
        \cline{2-3}
        & 6.67e-8 & if memory $\le$ 4096MB \\
        \cline{2-3}
        & 8.83e-8 & if memory $\le$ 5120MB \\
        \cline{2-3}
        & 1e-7 & if memory $\le$ 6144MB \\
        \cline{2-3}
        & 1.167e-7 & if memory $\le$ 7168MB \\
        \cline{2-3}
        & 1.333e-7 & if memory $\le$ 8192MB \\
        \cline{2-3}
        & 1.5e-7 & if memory $\le$ 9216MB \\
        \cline{2-3}
        & 1.667e-7 & if memory $\le$ 10240MB \\
        \hline
        $price\_cpu_{inst}$ & \multicolumn{2}{c}{3.02e-14} \\
        \hline
        \multirow{13}{*}{$price\_gpu_{cons}$}& 6.3e-9 & if memory $\le$ 128MB \\
        \cline{2-3}
        & 2.49e-8 & if memory $\le$ 512MB \\
        \cline{2-3}
        & 5.01e-8 & if memory $\le$ 1024MB \\
        \cline{2-3}
        & 7.5e-8 & if memory $\le$ 1536MB \\
        \cline{2-3}
        & 9.99e-8 & if memory $\le$ 2048MB \\
        \cline{2-3}
        & 1.5e-7 & if memory $\le$ 3072MB \\
        \cline{2-3}
        & 2.001e-7 & if memory $\le$ 4096MB \\
        \cline{2-3}
        & 2.499e-7 & if memory $\le$ 5120MB \\
        \cline{2-3}
        & 3e-7 & if memory $\le$ 6144MB  \\
        \cline{2-3}
        & 3.501e-7 & if memory $\le$ 7168MB \\
        \cline{2-3}
        & 3.999e-7 & if memory $\le$ 8192MB \\
        \cline{2-3}
        & 4.5e-7 & if memory $\le$ 9216MB \\
        \cline{2-3}
        & 5.001e-7 & if memory $\le$ 10240MB \\
        \hline
        $price\_gpu_{inst}$ & \multicolumn{2}{c}{9.06e-14} \\
        \bottomrule
    \end{tabular}
    \vspace{-0.2cm}
    \caption{Settings of the pricing parameters in Eq.(\ref{equation:price}).}
    \label{tab:prices}
    \vspace{-0.5cm}
\end{table}

\setcounter{table}{0}   
\setcounter{figure}{0}

\section{Details of Experiments}
\label{appendix:exp_impl}
\subsection{Training and Testing Sets}
For training sets, we select 2,760 and 480 samples from OpenCATP for sequential and non-sequential planning, respectively. As for testing sets, we select 720 and 480 samples from various tasks which are not present in the training sets for the evaluation of sequential and non-sequential planning, respectively.

\subsection{Implementation of CATP-LLM}
\label{appendix:impl_catpllm}
The details of implementing CATP-LLM on OpenCATP platform in the experiments are described as follows. 

\noindent \textbf{Details of TPL.} For TPL, we assign each tool in OpenCATP with a unique tool token and a dependency token. The dependency token of a tool means accepting the outputs provided by this tool as inputs, as depicted in Figure~\ref{fig:tpl_demo}. We also introduce a special dependency token $\langle task \rangle$ indicating accepting the task data as inputs. Note that in our experiments, we focus on resource dependencies, \ie, input-output dependencies. However, the concepts of TPL can be easily extended to other types of dependencies, such as order dependencies where there exist strict execution orders between tools. 

\noindent \textbf{Details of CAORL}. As for CAORL, we use the execution prices and task performance defined in $\S$\ref{appendix:evaluate_metrics} to calculate the execution costs and plan performance in the reward function. Besides, to implement the context augmentation scheme, we need to derive the cost attributes of each tool. To achieve this, we categorize the input data in OpenCATP into $k=4$ size levels  using k-means clustering, with the optimal number of clusters determined by the elbow method. We then profile the execution time and CPU/GPU memory consumption of each tool across varying input sizes, following the method outlined in $\S$\ref{subsubsec:context_augmentation}. Based on these statistics, we calculate the overall costs for each tool using the execution prices defined in Eq.(\ref{equation:price}), which ultimately yield the tool cost attributes.

\noindent \textbf{Details of fine-tuning.} We apply the data generation method described in $\S$\ref{subsec:caorl} to create 1,200 sequential tool plans and 780 non-sequential tool plans as training data. We then fine-tune CATP-LLM using Llama2-7B as the default backbone with LoRA rank 64. The total number of trainable parameters, including LoRA weights and learning tokens in TPL, are less than 1M, accounting for no more than 0.2\% of the total parameters. Besides, it only takes CATP-LLM 9.5h and 6.2h to converge for 2 epochs in the sequential and non-sequential scenarios, when fine-tuned over a single 32GB V100 GPU. Hence, the fine-tuning overhead of CATP-LLM is small. 

\subsection{Implementation of Baselines}
\noindent \textbf{Zero-shot learning.} For zero-shot learning, we design two types of prompts for sequential planning and non-sequential planning, respectively. For sequential planning, we prompt the LLM to produce a tool sequence that can be  executed sequentially to solve the target tasks. As for non-sequential planning, we instruct the LLM to generate a tool plan following the format similar to our TPL. This is because we find the LLM achieves poor performance in non-sequential planning without an appropriate format to generate non-sequential plans.
Note that we also incorporate the average tool cost information into the prompts to instruct the LLM to create low-cost tool plans. The average tool cost information is derived from offline profiling results described in $\S$\ref{appendix:impl_catpllm}.
The prompts for zero-shot learning are shown below.
\begin{lstlisting}
[Prompt for Sequential Planning]

# Instructions

 You need to act as a policy model that, given a task and a set of tools, determines the sequence of tools that can be executed sequentially to solve the task. Your goal is to optimize the task performance while minimizing the execution costs.

# Tool Information

 The information of each tool is provided as follows:
 
 Object Detection: This tool identifies the names of objects in an image. It is generally used for object identification in the iamges. The input and output types of this tool are image and text, respectively. It can accept inputs from tools ''Image Super Resolution'', ''Colorization''', ''Image Deblurring'', or ''Image Denoising''.
    
 Costs of Object Detection: On average, this tool takes about 175.73 milliseconds to run. It consumes approximately 352.19 MB of CPU memory and 449.37 MB of GPU memory.
    
 Image Deblurring: This tool can enhance the clarity of blurry images. It can be used for tasks that require improving image quality. Both the input and output types are images. It can accept inputs from tools ''Image Super Resolution'', ''Colorization'', or ''Image Denoising''.
    
 Costs of Image Deblurring: On average, this tool takes about 667.42 milliseconds to run. It consumes approximately 444.91 MB of CPU memory and 3498.11 MB of GPU memory.

 [...]

# Response Format

 Provide a response in the similar format according to the following example: ''Tool1, Tool2, Tool3''.

\end{lstlisting}
\begin{lstlisting}
[Prompt for Non-Sequential Planning]

# Instructions

 You need to act as a policy model that, given a task and a set of tools, determines the execution plan of tools that can be executed in sequential or parallel to solve the task. Your goal is to generate the tool plans that can optimize the task performance while minimizing the execution costs. 

# Tool Information

 [Same as Sequential Planning]

# Response Format

 Provide a response in the similar format according to the following example: ['Tool1', ['Task'], 'Tool2', ['Tool1'], 'Tool3', ['Tool1']]. The meaning of this format is that the input data of Tool2 comes from the outputs of Tool1. Besides, ['Task'] means that Tool1 depends on the input data provided by the task. Please generate plans strictly according to this format.

\end{lstlisting}

\noindent\textbf{Few-shot learning.} The few-shot learning share similar prompts with zero-shot learning, except that we handcraft several high-quality demonstrations as in-context examples to augment the LLM for plan generation. 
The prompts for few-shot learning are shown below.
\begin{lstlisting}
[Prompt for Sequential Planning]

# Instructions

 [Same as Zero-Shot]

# Tool Information

 [Same as Zero-Shot]

# In-Context Example

 Task: Given a low-resolution, noisy, blurry gray image, how to return the regular image step by step?

 Plan: Image Super Resolution, Image Denoising, Image Deblurring, Colorization

 [...]

# Response Format

 [Same as Zero-Shot]

\end{lstlisting}
\begin{lstlisting}
[Prompt for Non-Sequential Planning]

# Instructions

 [Same as Zero-Shot]

# Tool Information

 [Same as Zero-Shot]

# In-Context Example

 Task: Given a low-resolution grayscale image, how to (1) return the regular image, and (2) return the class label of the image in English?
 
 Plan: ['Colorization', ['Task'], 'Image Super Resolution', ['Colorization'], 'Image Classification', ['Colorization']]

 [...]

# Response Format

 [Same as Zero-Shot]

\end{lstlisting}

\noindent\textbf{HuggingGPT.} We reuse the prompt of HuggingGPT~\cite{shen2024hugginggpt} for planning in our experiments. Besides, we have made some small modifications on the prompt by providing the tool cost information of each tool.

\noindent\textbf{HYDRA.} We reuse the pipeline and the prompt of HYDRA\cite{ke2024hydra} in our experiments. To ensure fairness, we restrict the tool set of HYDRA to those available in OpenCATP. What's more, similar to HuggingGPT, we have also added the tool cost information in the prompt of HYDRA.

\noindent\textbf{Instruction fine-tuning (IFT) and RLTF.} We adapt our TPL and apply it on IFT and RLTF, as we find that they perform poorly especially in non-sequential planning if generating tool plans in natural language. 

Note that the engines of GPT-3.5 and GPT-4 used for all prompt-based methods are gpt-3.5-turbo-0125 and gpt-4-turbo, respectively.

\section{Evaluate CATP-LLM on General Benchmark}

\begin{table}[t]
    \centering
    \scriptsize
    \setlength{\tabcolsep}{0pt}
    \begin{tabular}{m{1.65cm}<{\centering} |  m{1.55cm}<{\centering}  
     m{2.3cm}<{\centering}   m{2.7cm}<{\centering}}
    \toprule
        \textbf{Method} & \textbf{Tool Planner} & \textbf{Mean Accuracy (\%)}  & \textbf{Average Number of Tools}  \\ 
        \midrule
        CoT GPT-3.5 & / & 78.31  & / \\ 
        CoT GPT-4 & / & 83.99  & / \\ 
        Chameleon [34] &  GPT-4  & 86.54 & 3.40 \\ 
        \midrule
        \multicolumn{4}{c}{Published Results (Above) $\blacktriangle$}\\
        \midrule
        \rowcolor{rcolor}
        {\textbf{CATP-LLM}}  & Llama2-7B & 85.86  & 2.41\\ 
        \bottomrule
    \end{tabular}
    \vspace{-0.2cm}
    \caption{Comparing CATP-LLM with other methods on ScienceQA. CoT GPT-3.5/4 answer questions without using tools.}
    \label{tab:scienceqa}
    \vspace{-0.25cm}
\end{table}

To validate the effectiveness of CATP-LLM on other benchmark except OpenCATP, we implement CATP-LLM on ScienceQA~\cite{lu2022learn}, a widely adopted multimodal benchmark  for multi-choice question answering. 
We reused the tools in ScienceQA developed by Chameleon~\cite{lu2024chameleon}, a few-shot learning-based tool planning method. As ScienceQA lacks implementation for cost measurement (\eg, memory consumption), we use the number of tools in each plan as the indicator for tool planning costs. As shown in Table~\ref{tab:scienceqa}, CATP-LLM outperforms CoT GPT-3.5 and GPT-4. Besides, CATP-LLM uses significantly fewer tools than Chameleon GPT-4 (a reduction of 29.11\%) while its accuracy is only marginally lower than that of Chameleon GPT-4 (a decline of 0.68\%). This highlights CATP-LLM's advantage in cost-aware tool planning to substantially reduce tool costs without sacrificing performance. \textit{This finding on ScienceQA is consistent with that on OpenCATP, which demonstrates the generalization ability of CATP-LLM.} 

\section{Evaluate CATP-LLM on Various Hardware Equipment}

\setcounter{table}{0}   

\renewcommand{\cellwidth}{1.3cm}
\renewcommand{\taskcellwidth}{1.2cm}
\renewcommand{\qopcellwidth}{1.1cm}

\begin{table}[t]
\centering
\tiny
\setlength{\tabcolsep}{1pt}
\begin{tabular}{ m{1cm}<{\centering}| c | m{\taskcellwidth}<{\centering}  m{\cellwidth}<{\centering} m{\cellwidth}<{\centering} m{\qopcellwidth}<{\centering}}
\hline
\textbf{Devices} & \textbf{Methods}  &  \textbf{Task Scores} \ensuremath{\uparrow} & \textbf{Exec. Price (\$)} \ensuremath{\downarrow} & \textbf{Runtime (s)} \ensuremath{\downarrow} & \textbf{QoP} \ensuremath{\uparrow}\\
  \hline
  \multirow{8}{*}{\makecell{\textbf{NVIDA RTX}\\\textbf{3090}}} & \multicolumn{5}{c}{\textit{Sequential Planning}} \\
  \cline{2-6}
  
    & HuggingGPT (GPT-4) & 0.665 & 0.097 & 1.108 & 0.246 \\
    & Hydra (GPT-4) & 0.603 & 0.061 & 0.654 & 0.247 \\
    & \cellcolor{rcolor} \textbf{CATP-LLM} (Llama2-7B) & \cellcolor{rcolor} 0.652 & \cellcolor{rcolor} 0.072 & \cellcolor{rcolor} 0.748 & \cellcolor{rcolor} \textbf{0.262} \\ 
  \cline{2-6}
    &  \multicolumn{5}{c}{\textit{Non-Sequential Planning}} \\
  \cline{2-6}
    &  HuggingGPT (GPT-4) & 0.419 & 0.159 & 2.099  & 0.069 \\
    &  Hydra (GPT-4) & 0.332 & 0.068 & 1.180 & 0.106 \\
    &  \cellcolor{rcolor} \textbf{CATP-LLM} (Llama2-7B) &\cellcolor{rcolor} 0.566 & \cellcolor{rcolor} 0.138 &\cellcolor{rcolor} 0.761 &\cellcolor{rcolor} \textbf{0.161} \\
    
  \hline
     \multirow{8}{*}{\makecell{\textbf{NVIDA RTX}\\\textbf{4090}}} & \multicolumn{5}{c}{\textit{Sequential Planning}} \\
  \cline{2-6}
  
    &  HuggingGPT (GPT-4)  & 0.667 & 0.087 & 1.228 & 0.257 \\
    &  Hydra (GPT-4) & 0.603 & 0.039 & 0.432 & 0.267 \\
    & \cellcolor{rcolor} \textbf{CATP-LLM} (Llama2-7B) & \cellcolor{rcolor} 0.651 & \cellcolor{rcolor} 0.045 & \cellcolor{rcolor} 0.487 & \cellcolor{rcolor}\textbf{0.286} \\ 
  \cline{2-6}
     & \multicolumn{5}{c}{\textit{Non-Sequential Planning}} \\
  \cline{2-6}
    &  HuggingGPT (GPT-4) & 0.419 & 0.103 & 1.413 & 0.118\\
    &  Hydra (GPT-4)  & 0.332 & 0.049 & 0.844 & 0.122 \\
    & \cellcolor{rcolor} \textbf{CATP-LLM} (Llama2-7B)  &\cellcolor{rcolor} 0.566 &\cellcolor{rcolor} 0.077 & \cellcolor{rcolor}0.463 &\cellcolor{rcolor} \textbf{0.215} \\
\hline
\end{tabular}
\vspace{-0.3cm}
\caption{Evaluation on real-world commercial GPUs. Arrow \ensuremath{\uparrow}/\ensuremath{\downarrow} means higher/lower is better.}
\label{tab:real_world}
\vspace{-0.3cm}
\end{table}

According to the cost definition in Equation (B.1), the cost measurement (\eg, runtime and memory consumption) can be affected by hardware equipment. 
To validate the applicability of CATP-LLM across diverse hardware configurations, we test the plans of CATP-LLM and top baselines on various real-world commercial GPUs. 
We report the performance of these methods in Table~\ref{tab:real_world}.
We can see that CATP-LLM strikes the better balance between task scores and execution costs, leading to the highest QoP. \textit{This demonstrates  the applicability of CATP-LLM in various hardware configurations.}

\end{document}